%% file: main_google_template.tex
\documentclass[11pt, a4paper, twocolumn, copyright, goog]{google}

\usepackage[authoryear, sort&compress, round]{natbib}

\usepackage{todonotes}
\usepackage{lipsum} 
\usepackage{multirow}
\usepackage{microtype}
\usepackage{xspace}

\usepackage{amsmath}      
\usepackage{amsfonts}     
\usepackage{algorithm}    
\usepackage{algorithmic}
\usepackage{float}        

\usepackage{graphicx}
\usepackage{subcaption}

\input{variables}


\uselogo{2} 

\title{\textit{LoRA-Squeeze}: Simple and Effective Post-Tuning and In-Tuning Compression of LoRA Modules}

\correspondingauthor{vulic@google.com}

\reportnumber{0001} 

\author[]{Ivan Vuli\'c}
\author[]{Adam Grycner}
\author[]{Quentin de Laroussilhe}
\author[]{Jonas Pfeiffer}

\affil[]{\thepa{ DeepMind}{}}

\begin{abstract}
Despite its huge number of variants, standard Low-Rank Adaptation (LoRA) is still a dominant technique for parameter-efficient fine-tuning (PEFT). Nonetheless, it faces persistent challenges, including the pre-selection of an optimal rank and rank-specific hyper-parameters, as well as the deployment complexity of heterogeneous-rank modules and more sophisticated LoRA derivatives. In this work, we introduce \textit{LoRA-Squeeze}, a simple and efficient methodology that aims to improve standard LoRA learning by \textit{changing LoRA module ranks either post-hoc or dynamically during training}. Our approach posits that it is better to first learn an expressive, higher-rank solution and then compress it, rather than learning a constrained, low-rank solution directly. The method involves fine-tuning with a deliberately high(er) source rank, reconstructing the full weight update matrix, and then using Randomized Singular Value Decomposition (RSVD) to create a new, compressed LoRA module at a lower target rank. Extensive experiments across 13 text-based and 10 vision-language tasks show that post-hoc compression often produces lower-rank adapters that outperform those trained directly at the target rank, especially if a small number of fine-tuning steps at the target rank is allowed. Moreover, a gradual, in-tuning rank annealing variant of \textit{LoRA-Squeeze} consistently achieves \textit{the best LoRA size-performance trade-off}. \textit{LoRA-Squeeze} decouples the training rank from the deployment rank, this way improving module efficiency, reusability, portability, and interoperability.

\end{abstract}

\begin{document}

\maketitle

\input{sections/01_introduction_icml}
\input{sections/02_rw_icml}
\input{sections/03_methodology_icml}

\input{sections/04_expsetup_icml}
\input{sections/05_results_icml}

\input{sections/06_conclusion_google}

\bibliography{main}

\newpage
\onecolumn
\appendix
\input{sections/xx_appendix_icml}

\end{document}

%% file: variables.tex
\newcommand{\rparagraph}[1]{\vspace{0.5mm}\noindent\textbf{#1.}}

\newcommand{\method}{\textsc{LoRA-Squeeze}\xspace}

\newcommand{\authortodo}[4]{%
    \expandafter\newcommand\csname #1\endcsname[1]{\todo[color=#2!40]{\textbf{#3:} \small ##1}}
    \expandafter\newcommand\csname #1inline\endcsname[1]{\todo[inline,color=#2!40]{\textbf{#3:} ##1}}
}

\authortodo{ivan}{green}{Ivan}{}
\authortodo{quentin}{cyan}{Quentin}{}
\authortodo{adam}{magenta}{Adam}{}
\authortodo{jonas}{yellow}{Jonas}{}

%% file: sections/01_introduction_icml.tex
\section{Introduction}
\label{sec:intro}
The ever-growing scale of large language models (LLMs) combined with the demand for their efficient adaptation has established Low-Rank Adaptation (LoRA) \citep{hu2022lora} as a leading, go-to Parameter-Efficient Fine-Tuning (PEFT) technique. LoRA operates on the hypothesis that the weight updates $\Delta W \in \mathbb{R}^{m \times n}$ of the large model with initial weights $W_0 \in \mathbb{R}^{m \times n}$ can be \textit{approximated} by a low-rank decomposition $\Delta W = AB$, where $A \in \mathbb{R}^{m \times r}$ and $B \in \mathbb{R}^{r \times n}$ are trainable parameters, and $r \ll \min(m, n)$ is the LoRA rank. Despite its widespread adoption and success, there are several critical challenges that persist in the practical applications of LoRA:
%

%

\noindent \textit{Challenge 1.} \textit{Having lower-rank LoRA-s that maintain performance} of higher-rank LoRA-s is one of the key desiderata to improve efficiency and module portability while simultaneously reducing latency and module storage requirements. 

\noindent \textit{Challenge 2.}\textit{ Maintaining architectural homogeneity of LoRA modules} (i.e., relying on same-rank LoRA-s) is often crucial for infrastructural simplicity and deployment. Put simply, LoRA-s of varying ranks and of heterogeneous decompositions may pose various challenges during deployment related to, e.g., inefficient batching and memory overheads during LoRA loading and unloading~\citep{slora,punica}.

\noindent \textit{Challenge 3.}\textit{ The selection of rank} $r$ typically must be done in advance of fine-tuning; the performance of a LoRA-adapted model is often highly sensitive to this choice~\citep{dylora}, and the optimal rank can vary across tasks and datasets of different complexities, and across different models and model sizes~\citep{zhang2023adalora}. 

\noindent \textit{Challenge 4.} Each rank might further necessitate its own \textit{hyperparameter sweeps} to determine, e.g., the best per-rank and per-LoRA-matrix learning rate and/or training schedule~\citep{hayou2024loraplus,schulman2025lora}. 

In this work, with pragmatism and deployment-savvy solutions as the principal drivers, we focus on improving \textit{efficiency}, \textit{portability}, and \textit{reusability} of standard, ubiquitous LoRA modules (see the rationale for this choice later in \S\ref{sec:rw}). To this end, we shed new light on and exploit tight connections between (i) standard LoRA modules, (ii) task arithmetic and task vectors~\citep{ansell-etal-2022-composable,ilharco2023editing}, and (iii) SVD decomposition, positing that the LoRA rank used for training should be \textit{decoupled} from the rank used for deployment, which has positive implications on the four challenges above. We introduce \textbf{\method}, a novel, computationally efficient methodology for changing the rank of standard LoRA modules. The changes can be applied either post-hoc after LoRA fine-tuning, or dynamically during their fine-tuning. The \method methodology relies on the two key principles:

\textit{Overparameterized Fine-Tuning:} LoRA is fine-tuned on a target task with a deliberately high rank. Higher-rank tuning is used during the whole fine-tuning process or during its parts. This allows the model to learn task adaptation within a less constrained, higher-dimensional space.

\textit{Compression:} Higher-rank LoRA modules can then be (i) gradually reduced/annealed during fine-tuning (i.e., dynamic \textit{in-tuning transformations}, dubbed \textit{In-Squeeze}, see Figure~\ref{fig:in-squeeze}) or (ii) post-hoc after fine-tuning (i.e., static \textit{post-tuning transformations}, dubbed \textit{Post-Squeeze}, see Figure~\ref{fig:post-squeeze}). Given that LoRA can be viewed as an approximation of the `full delta' $\Delta W$, LoRA matrices are first reconstructed into the `full model space' yielding the $\Delta W$ representation, and then compressed to a desired, lower rank using a very efficient implementation of Randomized Singular Value Decomposition (RSVD) \citep{halko2011finding}.

The two principles have several positive implications on the efficiency, portability and reusability aspects of LoRA modules. For instance, as our results over 13 standard text-based and 10 vision-text tasks demonstrate, \method helps unify LoRA structures after fine-tuning in cases when they were fine-tuned with different ranks without the need for retraining with other ranks, having positive impact on architectural homogeneity. Further, it is not required to conduct hyperparameter search per each rank separately; it is possible to find a good setup for a subset of higher-order modules and create same- or even higher-quality lower-rank LoRA-s post-hoc, without any fine-tuning. Finally, we empirically validate that the gradual in-tuning \method procedure enables fine-tuning LoRA modules with a better trade-off between size and task performance than direct fine-tuning with a chosen rank.

Put simply, our work shows that for LoRA it is often better to first learn a more expressive, higher-rank solution and then compress it, rather than attempting to learn a fixed low-rank solution from the outset. This paradigm not only offers immediate practical benefits, but also deepens our understanding of the mechanisms underlying parameter-efficient adaptation with LoRA-s.

%% file: sections/02_rw_icml.tex
\section{Background and Related Work}
\label{sec:rw}

\noindent \textbf{On SVD and LoRA.}
Recent research has increasingly leveraged SVD—a fundamental technique for matrix factorization and optimal low-rank approximation—to enhance the 1) initialization, 2) structure, and 3) optimization of LoRA.

First, traditional LoRA initializes one matrix randomly and the other with zeros, which may not effectively leverage the structure of the pretrained weights $W_0$. SVD provides a principled approach to initialization by decomposing the original weight matrix. PiSSA (Principal Singular Value Adaptation)~\citep{meng2025pissa} and LoRA-Null~\citep{tang2025loranulll} utilize this by initializing the LoRA adapters with the principal components (i.e., those associated with the largest singular values) of the pretrained weights. This strategy aims to capture the most salient information from the base model, often leading to faster convergence. Furthermore, approaches like MiLoRA (Minor Low-Rank Adaptation) \citep{zhang2024milora} suggest focusing adaptation on the minor smallest singular values/components, arguing that the principal components already capture essential knowledge. SVD-informed initialization strategies have also been extended to more complex architectures; e.g., GOAT~\citep{fan2025makeloragreatagain} introduces an SVD-structured Mixture-of-Experts framework where different experts are initialized with distinct SVD segments. 

Second, SVD has also been integrated directly into the LoRA structure to achieve extreme parameter efficiency or inspire novel architectures. LoRA-XS~\citep{bałazy2025loraxs} achieves drastic parameter reduction by leveraging the SVD of the original weight matrix to create frozen low-rank matrices (U and V). A very small, trainable matrix is then inserted between them, significantly reducing the number of trainable parameters by only training the interaction between the frozen singular vectors.

Finally, SVD provides a mathematical basis for analyzing and optimizing rank allocation dynamically. AdaLoRA~\citep{zhang2023adalora} directly addresses this by parameterizing the low-rank updates in an SVD form. It iteratively estimates the importance of weight updates  based on an importance score derived from an SVD-based parameterization of the weight updates, and then prunes the least significant ones. This dynamically optimizes the rank distribution across the model, allocating more capacity to critical layers while reducing redundancy in others. While AdaLoRA addresses the same fundamental problem as our work, the sub-optimality of a uniform, prespecified rank allocation across all layers and modules, it introduces extra complexity into the training loop; it requires importance scoring and budget scheduling. Moreover, by design it breaks the homogeneity of LoRA structures~\citep{zhou-etal-2024-autopeft} which might have detrimental impact on infrastructural simplicity of deployment and serving LoRA modules~\citep{slora}. \method offers a simple, more flexible alternative without any training and serving overhead.

Another very related line of work coupling SVD and LoRA focuses on finding lower-ranked LoRA subspaces for a set of $N$ independently trained (higher-ranked) LoRA-s via spectral decomposition techniques~\citep{kaushik2025iccv,lora-core}, formalizing the idea of the existence of a shared lower-ranked subspace. In turn, this work combined with the universal weight subspace hypothesis~\cite{universalweight} hints that the `intrinsic dimension' of the delta/update is lower than the actual chosen rank, which sets theoretical foundations for \method. Here, we focus on a novel setup of retaining or improving performance via spectral decomposition for a single LoRA, rather than finding a shared subspace for $N$ LoRA-s. 

More generally, beyond the methods that rely directly on SVD, the number of module variants derived from the standard LoRA architecture, and again focusing on improved initialization, structure or optimization, is immense~\citep[among others]{li2024vblora,liu2024dora,kopiczko2024vera,li2025unilora,wu2024mole,tian2024hydralora,yang2024corda}. Despite all the variants, the standard LoRA design still remains a dominant paradigm due to its architectural and serving simplicity. More sophisticated variants often imply intricate interventions into the training process (e.g., customized matrix decompositions as with LoRA-XS, VeRA or DoRA), or creation of customized large base models (e.g., residual models as with PiSSA and MiLoRA), or yield non-homogeneous LoRA-s across the base model incurring deployment and serving difficulties (AdaLoRA). Therefore, for the variety of practical reasons including its wide adoption, in this work we deliberately focus on the setup which relies on the standard LoRA architecture, and where such modules are directly tied to standard LLM checkpoints. The extension of the \method principles to more sophisticated LoRA architectures is left for future research.

Within this setup, \method thus aims to (1) reduce trainable parameters of standard LoRA-s via in-tuning or post-tuning rank reduction, (2) initialize lower-rank LoRA-s with SVD of weights that originated from higher-rank LoRA training, and (3) optimize resource allocation offline or without any significant training overhead.  

%% file: sections/03_methodology_icml.tex
\begin{figure}[t!]
    \centering 
        \includegraphics[width=0.99\linewidth]{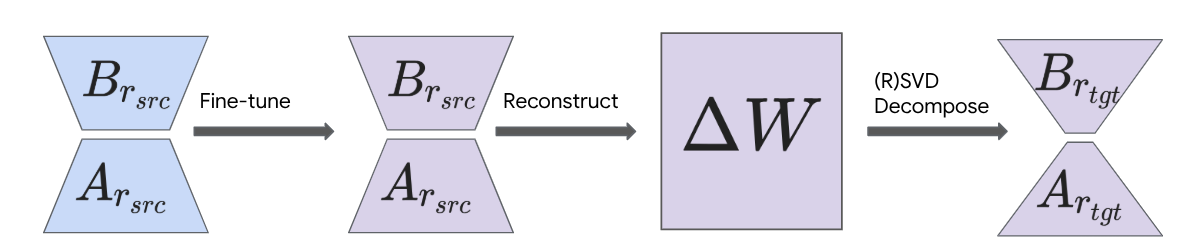}
        \caption{\method \textit{after} fine-tuning (\textit{Post-Squeeze}). We fine-tune with a LoRA with a higher, `source' LoRA rank $r_{src}$ and then transform it to a lower, `target' LoRA rank $r_{tgt}$.}
        \label{fig:post-squeeze}
\end{figure}

\begin{figure*}[t!]
    \centering 
        \includegraphics[width=0.99\linewidth]{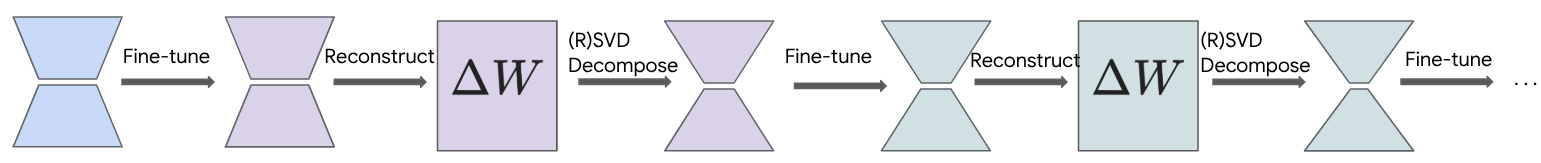}
        \caption{\method \textit{during} fine-tuning (\textit{In-Squeeze}); we can gradually anneal the LoRA rank during fine-tuning by reconstructing the full delta $\Delta W$ from the current LoRA, decompose it to a lower-rank LoRA via Randomized SVD and continue fine-tuning with a lower-rank. It repeats the main \textit{Post-Squeeze} steps (Figure~\ref{fig:post-squeeze}) multiple times during fine-tuning using a predetermined annealing scheme.}
        \label{fig:in-squeeze}
\end{figure*}

\section{Methodology}

The core of \method is a transformation that leverages Randomized SVD to compress a fine-tuned LoRA module from a high source rank to a lower target rank, as illustrated in Figure~\ref{fig:post-squeeze}. 

\noindent \textbf{Preliminaries I: LoRA as a Low-Rank Task Vector/Tensor Approximation.}
Let us assume a pretrained LLM parameterized by a set of weight matrices $\{W_{0,l}\}$, with $l=1,\ldots,L$, and $L$ is the number of LLM layers.\footnote{For notational simplicity, we will drop indexing over layers.} After fully fine-tuning for a task $T$, each weight matrix $W_0$ is updated to a new state of weights $W_T$. The change induced by the fine-tuning process can be captured by the \textit{task vector} (or, more accurately, the \textit{task tensor}), defined as the element-wise difference between the fine-tuned and pretrained weights~\citep{ilharco2023editing}: $\Delta W = W_T - W_0$.

The task tensor $\Delta W$ resides in the same high-dimensional space as the original weights and encapsulates the knowledge required to adapt the model for task $T$. LoRA is predicated on the empirical observation that this task vector $\Delta W$ typically has a low \textit{intrinsic rank}~\citep{hu2022lora, aghajanyan2021intrinsic}. Consequently, instead of learning the full dense matrix $\Delta W$, LoRA approximates it with a low-rank factorization. Thus, LoRA can be understood as a method for learning a parameter-efficient, low-rank approximation of the full-space task vector, where higher-rank approximations have more capacity, and can therefore fit a better approximation of the task vector.

\noindent \textbf{Preliminaries II: Randomized SVD.}
The use of low-rank decomposition, particularly Singular Value Decomposition (SVD), for compressing neural networks has a long history~\citep{cohen2021lp}. For any given matrix, truncated SVD provides the optimal low-rank approximation in the sense of minimizing the Frobenius norm of the difference, making it a principled choice for weight matrix compression. To efficiently handle the potentially large dimensions of $\Delta W$ for large LLMs, we apply a rank-$r$ truncated Randomized SVD (RSVD)~\citep{halko2011finding}.\footnote{We refer the reader to this blog post for an informative overview of RSVD: \url{https://gregorygundersen.com/blog/2019/01/17/randomized-svd/}.} As the standard SVD, RSVD decomposition of some matrix $M \in \mathbb{R}^{m \times n}$ yields three matrices corresponding to the top $r$ singular components: $\text{RSVD}_{r}(M) \rightarrow U, \Sigma, V^T$. Here, $U \in \mathbb{R}^{m \times r}$ contains the top left singular vectors, $\Sigma \in \mathbb{R}^{r \times r}$ is a diagonal matrix of the top singular values, and $V^T \in \mathbb{R}^{r \times n}$ contains the top right singular vectors.

\subsection{\method}

\renewcommand{\algorithmicrequire}{\textbf{Input:}}
\renewcommand{\algorithmicensure}{\textbf{Output:}}
\renewcommand{\algorithmiccomment}[1]{\hfill $\triangleright$ #1}
\newcommand{\RETURN}{\STATE \textbf{return}\xspace}

\begin{algorithm}[t!]
\caption{Randomized SVD for LoRA Creation}
\label{alg:rand_svd_lora}
{\small
\begin{algorithmic}[1] 
    \REQUIRE Input matrix $W \in \mathbb{R}^{m \times n}$, target rank $r$, over-sampling hyper-parameter $k_o$, number of power iterations $k_q$.
    \ENSURE LoRA matrices $A \in \mathbb{R}^{m \times r}$ and $B \in \mathbb{R}^{r \times n}$ such that $W \approx AB$.
    
    \STATE $r' \leftarrow r + k_o$
    \STATE Draw a random Gaussian matrix $\Omega \in \mathbb{R}^{n \times r'}$
    \STATE $Y \leftarrow W\Omega$
    \FOR{$i = 1 \to k_q$}
        \STATE $Q, \_ \leftarrow \text{qr\_decomposition}(Y)$
        \STATE $Y^* \leftarrow W^T Q$
        \STATE $Q^*, \_ \leftarrow \text{qr\_decomposition}(Y^*)$
        \STATE $Y \leftarrow W Q^*$
    \ENDFOR
    \STATE $Q, \_ \leftarrow \text{qr\_decomposition}(Y)$
    \STATE $D \leftarrow Q^T W$
    \STATE $\tilde{U}, S, V^T \leftarrow \text{svd\_decomposition}(D)$
    \STATE $U \leftarrow Q \tilde{U}$
    \STATE $U_r \leftarrow U[:, 1:r]$
    \STATE $S_r \leftarrow S[1:r]$
    \STATE $V_r^T \leftarrow V^T[1:r, :]$
    \STATE $\Sigma_r^{1/2} \leftarrow \text{diag}(\sqrt{S_r})$
    \STATE $A \leftarrow U_r \Sigma_r^{1/2}$
    \STATE $B \leftarrow \Sigma_r^{1/2} V_r^T$
    \RETURN $A, B$
\end{algorithmic}
    
    
    
}%
\end{algorithm}
\begin{algorithm}[t!]
\caption{\em Memory-Efficient \method}
\label{alg:efficient_squeeze}
{\small
\begin{algorithmic}[1]
\REQUIRE LoRA matrices $A_{src} \in \mathbb{R}^{m \times r_{src}}$, $B_{src} \in \mathbb{R}^{r_{src} \times n}$, target rank $r_{tgt}$
\ENSURE Pruned LoRA matrices $A_{tgt} \in \mathbb{R}^{m \times r_{tgt}}$, $B_{tgt} \in \mathbb{R}^{r_{tgt} \times n}$

\STATE \textbf{Step 1: Orthogonalize bases}
\STATE $Q_A, R_A \leftarrow \text{qr\_decomposition}(A)$ 
\STATE $Q_B, R_B \leftarrow \text{qr\_decomposition}(B^\top)$

\STATE \textbf{Step 2: Compute core interaction matrix}
\STATE $M \leftarrow R_A R_B^\top$ \COMMENT{Dense matrix of size $r_{src} \times r_{src}$}

\STATE \textbf{Step 3: Full SVD or RSVD}
\STATE $U_M, S_M, V_M^\top \leftarrow \text{(r)svd\_decomposition}(M)$ 

\STATE \textbf{Step 4: Truncate rank}
\STATE $U_r \gets U_M[:, 1:r_{tgt}]$
\STATE $S_r \gets S_M[1:r_{tgt}]$
\STATE $V_r^T \gets V_M^T[1:r_{tgt}, :]$

\STATE $\Sigma_r^{1/2} \gets \text{diag}(\sqrt{S_r})$

\STATE \textbf{Step 5: Reconstruct target-rank A and B}
\STATE $A_{tgt} \leftarrow Q_A U_r \Sigma_r^{1/2}$
\STATE $B_{tgt} \leftarrow \Sigma_r^{1/2} V_r^\top Q_B^\top$

\RETURN $A_{tgt}, B_{tgt}$
\end{algorithmic}
}%
\end{algorithm}

The main `building block' of \method operates on a LoRA module that has already been (partially or fully) fine-tuned with a relatively high source rank, $r_{src}$, and transforms it into a new module with an arbitrary, typically lower, target rank, $r_{tgt}$. This process, illustrated in Figure~\ref{fig:post-squeeze}, proceeds in three main steps:

\noindent \textit{Step 1: (Higher-Rank) Fine-Tuning:} First, a standard LoRA fine-tuning process is executed for a given task, but with a source rank $r_{src}$ chosen to be larger than the anticipated final deployment rank. This step yields the trained LoRA matrices, $A_{src} \in \mathbb{R}^{m \times r_{src}}$ and $B_{src} \in \mathbb{R}^{r_{src} \times n}$.

\noindent \textit{Step 2: Full Task Vector Reconstruction:} The high-rank LoRA matrices are multiplied to reconstruct the low-rank approximation of the task vector in the full parameter space. This results in the delta matrix $\Delta W_{src} = A_{src} \cdot B_{src}$. $\Delta W_{src}$ has the same dimensions as the original weight matrix $W_0$ but is constrained to have a rank of at most $r_{src}$.

\noindent \textit{Step 3: RSVD for LoRA Creation:} Finally, the components from the RSVD applied to $\Delta W_{src}$ are used to construct the new, compressed LoRA matrices, $A_{tgt}$ and $B_{tgt}$, for the arbitrary target rank $r_{tgt}$. The full procedure is summarized in Alg.~\ref{alg:rand_svd_lora}, where Lines 1-13 depict the step-by-step formulation of RSVD; SVD in Line 12 defines standard, full SVD.\footnote{To balance the magnitudes of the two resulting matrices, the singular values are distributed between them (L18-19 of Alg~\ref{alg:rand_svd_lora}).} 

The resulting matrices, $A_{tgt} \in \mathbb{R}^{m \times r_{tgt}}$ and $B_{tgt} \in \mathbb{R}^{r_{tgt} \times n}$, form a new LoRA of rank $r_{tgt}$. From the task arithmetic perspective, they constitute a new $r_{tgt}$-rank approximation of the task vector.

\noindent{\textit{Post-Squeeze} and \textit{In-Squeeze}.}
The process described above may be done only once and offline, after fine-tuning a LoRA with $r_{src}$: we refer to it as \textit{Post-Squeeze}. However, it also possible to iteratively repeat it during fine-tuning, as a \textit{gradual rank annealing} process. We refer to this variant as \textit{In-Squeeze}, illustrated in Figure~\ref{fig:in-squeeze}. A special case of \textit{In-Squeeze}, labeled \textit{Cont-Squeeze}, \textit{continues fine-tuning} at the final target rank, without any subsequent iterations (i.e., it does only a single iteration). Allowing additional fine-tuning after the rank reduction allows for recovering task performance when a lot of information gets discarded via RSVD decomposition (e.g., in cases when $r_{tgt} \ll r_{src}$).

\noindent \textbf{Memory-Efficient \method.}
A major constraint of the proposed \method method lies in Step 2 of the procedure (see Figure~\ref{fig:post-squeeze}) which requires full task vector reconstruction. This effectively creates a set of parameters of the size of the original LLM, which can be memory-inefficient for large-scale LLMs. We thus propose a memory-efficient variant of \method whose memory requirements are not bound by the full rank, but rather than on rank $r_{src}$, and can be used for \textit{Post-Squeeze}, \textit{In-Squeeze}, and \textit{Cont-Squeeze}. It is summarized in Alg.~\ref{alg:efficient_squeeze}.

The main idea is identifying the least-contributing dimensions without the need to reconstruct $\Delta W$ at the full rank. The matrix $M$ is valid for our purpose because QR decompositions of $A_{src}$ and $B_{src}$ matrices create orthogonal matrices $Q_A$ and $Q_B$ which only do rotation. They do not change the length (magnitude) or `importance' of the vectors they multiply (which are placed in the $R$ matrices). This means that the information about `how important a dimension is' (the singular values) is preserved perfectly inside the smaller interaction matrix $M$. A formal derivation describing the relationship between the standard and memory-efficient \method is provided in Appendix~\ref{app:derivation}.

Memory-efficient \method does not require the full $m\times n$ $\Delta W$ interaction matrix on which Full or Randomized SVD operates, but it is based on a smaller $r_{src} \times r_{src}$ matrix (see line 5 of Alg.~\ref{alg:efficient_squeeze}) which gets decomposed by SVD. In consequence, this creates a smaller `$r_{src}$-bound' rather than `$\{m,n\}$-bound' memory footprint. Additional insights into (approximated) computational complexity of different compositions within \method are in Appendix~\ref{app:complexity}.

%% file: sections/04_expsetup_icml.tex
\section{Experimental Setup}
\label{s:exp}

We conduct a comprehensive set of experiments across multiple model sizes, task domains and tasks of varying complexity, as well as over various rank configurations. The experiments are centered around the \textit{Gemma 3} family of models~\citep{gemma3}; the primary model for development and the majority of experiments is the instruction-tuned \textit{Gemma 3 4B IT} variant. We also conduct supplementary experiments with the smaller (text-only) \textit{Gemma 3 1B IT} and the larger \textit{Gemma 3 12B IT} variants.\footnote{Preliminary experiments on pretrained (PT) model variants such as Gemma 3 1B/4B/12B PT resulted in very similar findings.} 




\noindent \textbf{Evaluation Tasks.} For the majority of our empirical analysis, we rely on a suite of 13 standard scoring-based, text-based evaluation tasks. These tasks were selected for their diversity, simplicity of evaluation, as well as for covering a range of natural language understanding capabilities. We also experiment with 10 well-established vision-language (VL) QA tasks. For all tasks, evaluation is run on the standard test splits where available, and on dev otherwise; Appendix~\ref{app:tasks} lists the tasks and the corresponding datasets.




\subsection{LoRA Fine-Tuning Protocols}
\label{ss:lora_protocols}
Following a standard practice~\citep{liu2023visual,dai2023instructblip}, in all experiments LoRA-s are applied exclusively to the weight matrices of the text-processing components of the Gemma 3 models. This means that for the multi-modal experiments, the vision encoder (SigLIP)~\citep{zhai2023sigmoid} and the multi-modal projection layers were kept frozen. The main baseline are LoRA-s trained directly at a target rank. 

\noindent \textbf{Rank Configurations.}
We trained baseline LoRA modules directly at a range of ranks: $r \in \{1, 2, 4, 8, 16, 32, 64, 128\}$. For \textit{Post-Squeeze}, we compressed them to various lower target ranks ($r_{tgt}$). For \textit{In-Squeeze}, we also validated a range of source and target rank configurations. 

\noindent \textbf{Learning Rate Selection.} 
For all direct LoRA fine-tuning experiments, we first determined the optimal learning rate for each LoRA rank and model size combination to ensure that our baseline comparisons were as strong as possible~\citep{schulman2025lora}. We used the Adafactor optimizer~\citep{shazeer2018adafactor} with a linear warmup of 1000 steps and no subsequent learning rate decay. The optimal learning rate was selected via a grid search over the set \{0.001, 0.003, 0.01, 0.03, 0.1\}. The finally selected learning rates are provided in Table~\ref{tab:lr_sweep_appendix} in Appendix~\ref{app:lr_sweep}. In another experiment, we also analyze how \method can be used to create strong lower-rank LoRA modules without running any hyperparameter sweeps for the target rank.

\noindent \textbf{RSVD Configuration.}
Following standard practices~\citep{halko2011finding}, we set the number of oversampling dimensions $k_o=10$ and perform two subspace iterations $k_q=2$; see Alg.~\ref{alg:rand_svd_lora}. These choices provide a strong trade-off between efficiency and approximation quality.\footnote{Given the stochastic nature of RSVD, we also ran it with multiple random seeds, but observed minimal variation in task performance. In our preliminary experiments we also verified that resorting to cheap and efficient RSVD instead of full SVD yielded minimal, if any, degradation in performance for our set of tasks.} \footnote{The computation cost of RSVD, even when used multiple times as in case of \textit{In-Squeeze}, is negligible compared to the other components of LoRA fine-tuning, and RSVD can be run on CPU.} The same RSVD config is used with the memory-efficient \method.

\captionsetup[subfigure]{margin={5mm,-2mm}}
\begin{figure*}[t!] 
    \centering 
        \begin{subfigure}[h]{0.32\textwidth}
        \centering
        \includegraphics[width=0.99\linewidth]{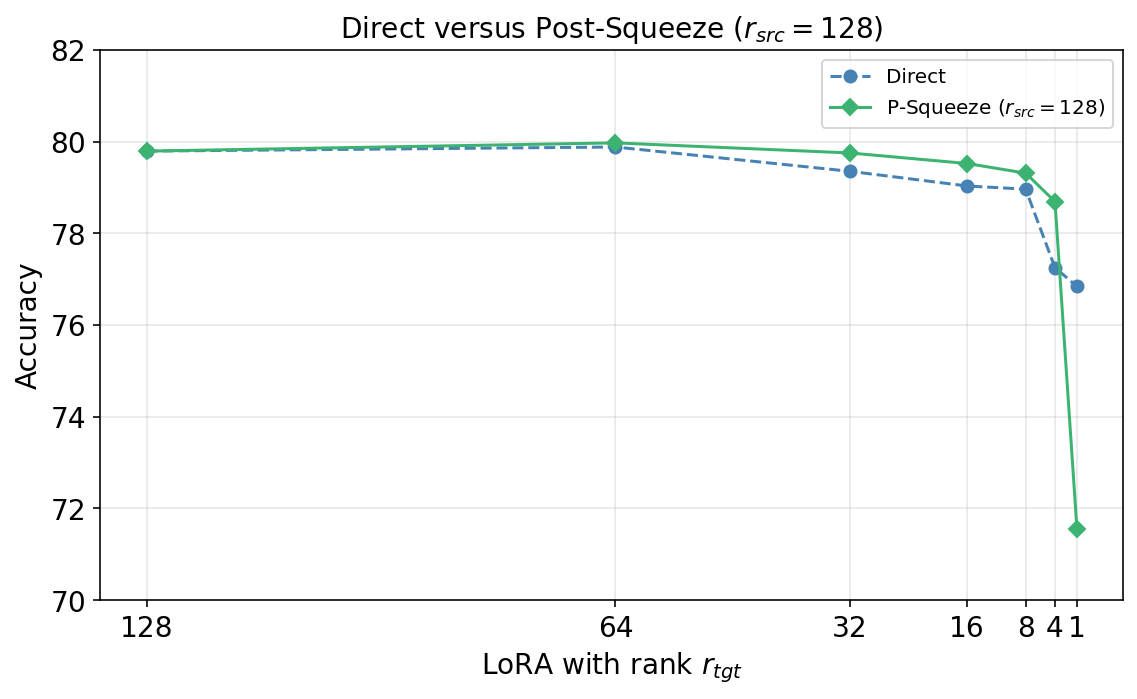}
        \caption{ANLI-r2}
        \label{fig:subopt_anli}
    \end{subfigure}
    \hfill 
    \begin{subfigure}[h]{0.32\textwidth}
        \centering
        \includegraphics[width=0.99\linewidth]{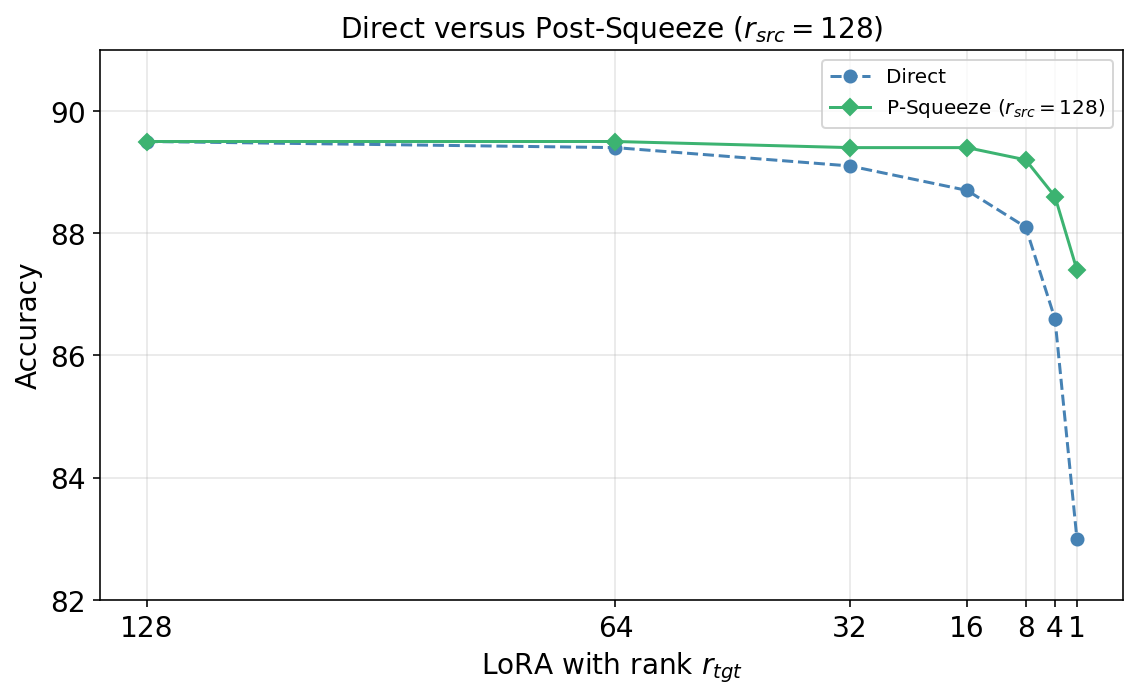}
        \caption{DROP}
        \label{fig:subopt_drop}
    \end{subfigure}
    \hfill
    \begin{subfigure}[h]{0.32\textwidth}
        \centering
        \includegraphics[width=0.99\linewidth]{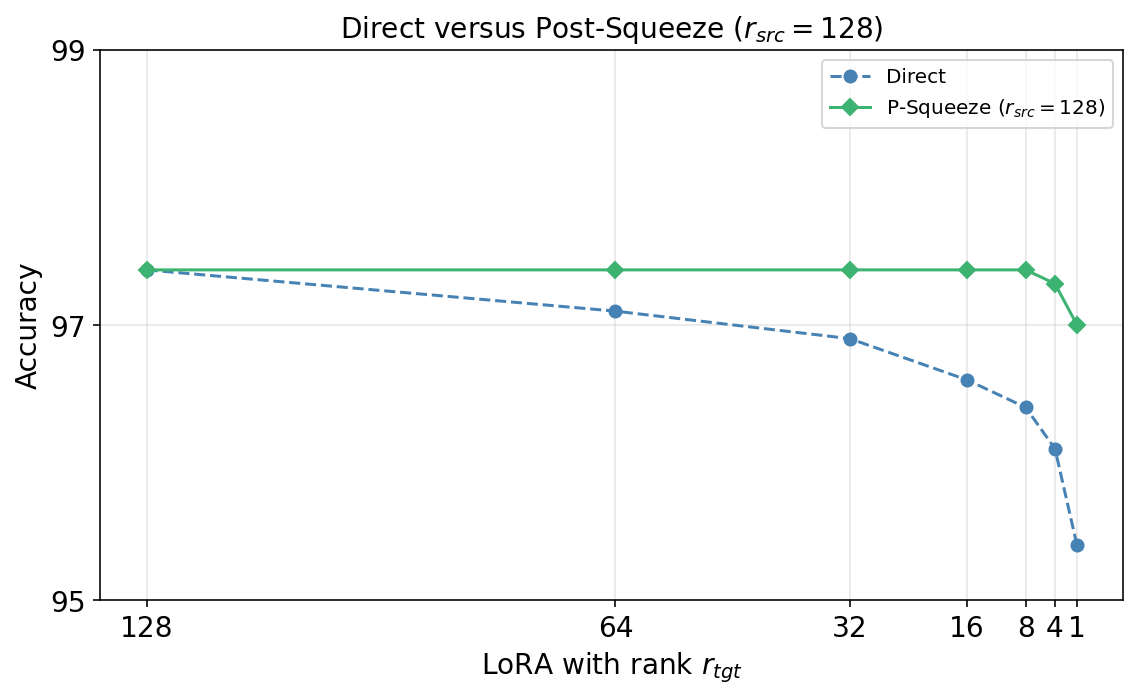}
        \caption{HellaSwag}
        \label{fig:subopt_hswag}
    \end{subfigure}

    \caption{Performance over 3 representative text-based tasks when we do hyperparameter search for the learning rate or LoRA-s only for the highest rank in the figures ($r_{src}=128$), and keep the same lr for direct fine-tuning at all the other (lower) ranks. A simple offline \textit{Post-Squeeze} method can bypass the hyperparameter search and yield better-performing LoRA-s without any fine-tuning at the lower ranks. Similar patterns are observed for the VL tasks; see the selection of plots in Figure~\ref{fig:subopts_vl} in Appendix~\ref{app:additional}. \textit{Remark:} For the higher results with $r_{tgt}$-rank LoRAs, where a learning rate sweep for $r_{tgt}$ was performed, we refer the reader later to Table~\ref{tab:finetuning_strategies_4b_text}.
    }
    \label{fig:suopts}
\end{figure*}


\noindent \textbf{Other Hyperparameters.}
In all experiments, we rely on a batch size of 8 and max sequence length is set to 1,024. To ensure convergence, models were fine-tuned for 10,000 steps on smaller datasets and 15,000 steps on larger datasets.

\noindent \textbf{\textit{In-Squeeze} Fine-Tuning Setup.}
\label{ss:in_tuning_exp}
For continued fine-tuning (i.e., the \textit{Cont-Squeeze} variant), directly trained as well as `post-squeezed' LoRA-s of the same rank $r_{tgt}$ are subjected to additional, short and inexpensive fine-tuning for 200 and 700 steps, using a 100-step warmup and the optimal learning rate previously determined for $r_{tgt}$-rank LoRA-s.

For the more general \textit{In-Squeeze} variant, fine-tuning starts at a high rank (e.g., 128) for a fraction of the total steps, after which LoRA gets squeezed to the next lower rank (e.g., 64). This process is repeated through subsequent lower ranks until the end rank (e.g., $r_{tgt}=1$). The total number of training steps is kept constant across all experiments for a fair comparison. We test two schemes for allocating the training budget across these stages. \textit{(1) Standard Scheme:} The total step budget is distributed among the rank stages proportionally to the rank value; e.g., in the setup where we anneal $128\rightarrow64\rightarrow32\rightarrow16\rightarrow8\rightarrow4\rightarrow2\rightarrow1$, the `rank-128 stage' receives $128 / (128+64+...+1)$ of the total steps; \textit{(2) Minimum Steps Scheme:} To ensure that lower ranks receive adequate training, each stage is first allocated a minimum of 200 steps. The remaining training budget is then distributed proportionally, as in the standard scheme.

\begin{figure*}[t!] 
    \centering 

    \begin{subfigure}[h]{0.49\textwidth}
        \centering
        \includegraphics[width=0.99\linewidth]{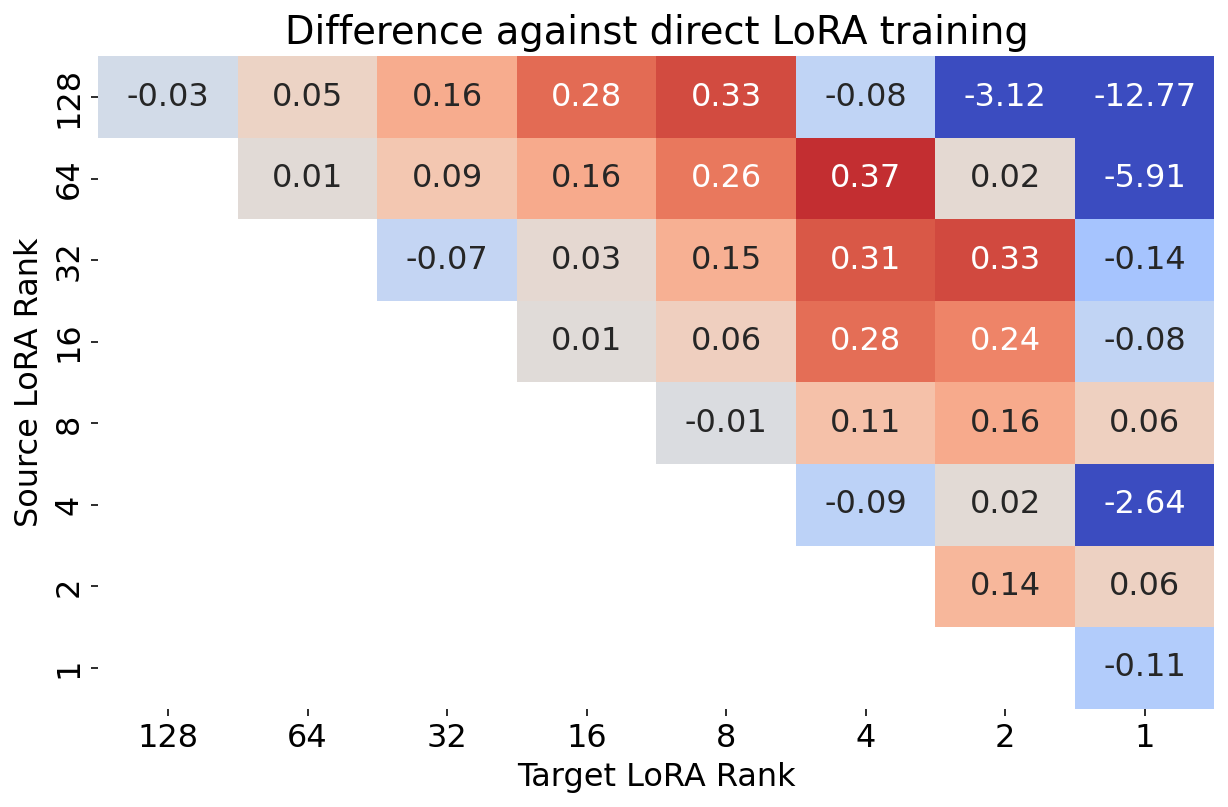}
        \caption{Gemma 3 4B IT.}
        \label{fig:hmap_4b}
    \end{subfigure}
    \begin{subfigure}[h]{0.49\textwidth}
        \centering
        \includegraphics[width=0.99\linewidth]{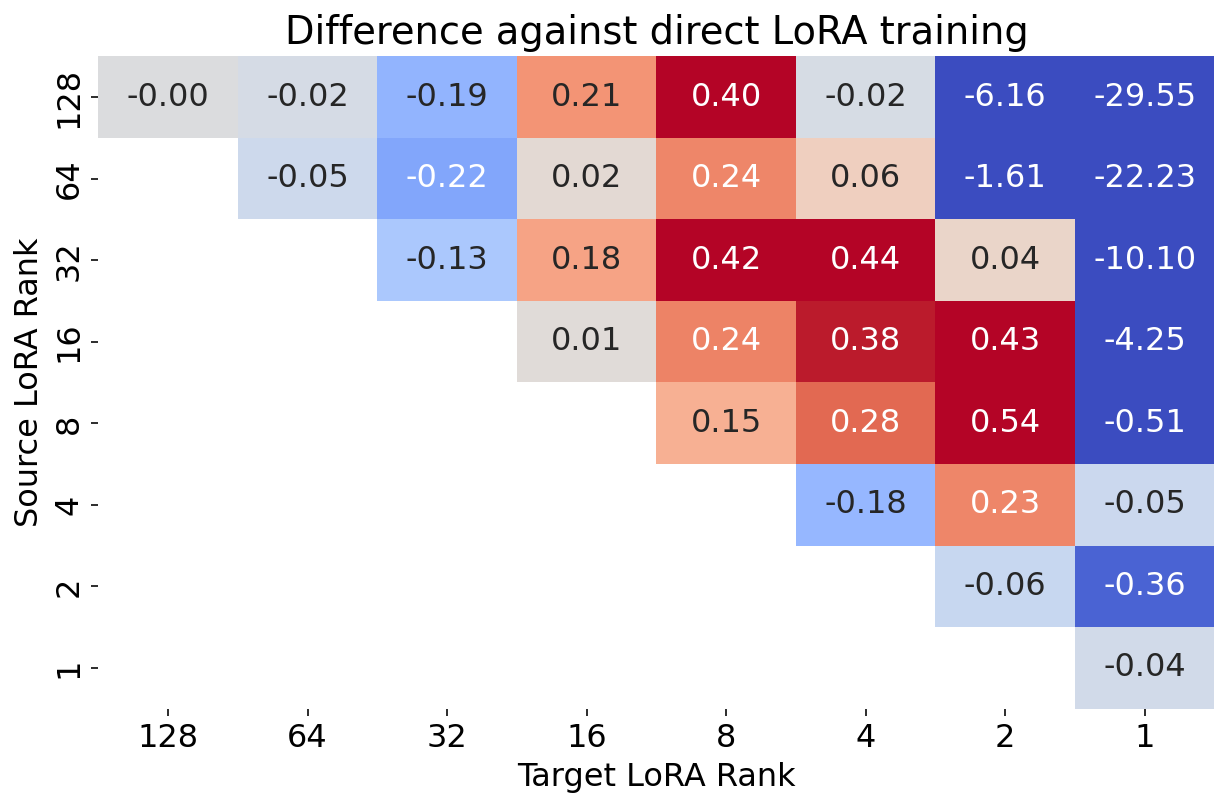}
        \caption{Gemma 3 1B IT.}
        \label{fig:hmap_1b}
    \end{subfigure}

    \caption{Performance difference heatmaps on text tasks for \textbf{(a)} Gemma 3 4B IT and (b) Gemma 3 1B IT. Each heatmap plots the average performance gain of Post-Squeeze from a given source rank $r_{src}$ (y-axis) to a target rank $r_{tgt}$ (x-axis), relative to a baseline LoRA module trained directly at $r_{tgt}$. Red cells indicate a positive gain, signifying that Post-Squeeze outperforms direct fine-tuning.}
    \label{fig:hmaps}
\end{figure*}

%% file: sections/05_results_icml.tex
\section{Results and Discussion}
\label{sec:results}

Our key experiments provide robust evidence that the \method methodology not only offers greater flexibility, but also often surpasses performance of standard (direct) LoRA fine-tuning, especially for the lowest ranks. 

\noindent \textbf{Reducing Hyperparameter Search.}
We first investigate a scenario where the aim is to obtain $r_{tgt}$-rank LoRA-s while reducing the number of per-rank hyperparameter sweeps (e.g., for the learning rate). Put simply, we conduct a hyperparameter sweep only for a single, source rank $r_{src}=128$, then create a series of lower-rank LoRA-s with \textit{Post-Squeeze}, and compare against directly fine-tuned $r_{tgt}$-rank LoRA-s using the same learning rate, which potentially might be suboptimal for the lower-rank LoRA-s. 

The results for three text tasks are in Figures~\ref{fig:subopt_anli}-\ref{fig:subopt_hswag}, and the results for two VL tasks are in Figures~\ref{fig:subopt_okvqa}-\ref{fig:subopt_tallyqa} in Appendix~\ref{app:additional}. 
%
%
They clearly show how (i) suboptimal learning rates can yield suboptimal results with direct fine-tuning of LoRA-s with different ranks (i.e., the practitioner thus indeed has to carefully fine-tune crucial hyper-parameters per LoRA rank), and how (ii) per-rank hparam sweep for a multitude of target ranks can be avoided with \method.

Nonetheless, in order to enable fair per-rank comparisons, all the remaining experiments rely on per-rank optimized learning rates; see again~\S\ref{ss:lora_protocols} and Appendix~\ref{app:lr_sweep}.


\noindent \textbf{Post-Squeeze versus Direct Fine-Tuning.}
 We now examine the performance difference between \textit{Post-Squeeze} and direct LoRA fine-tuning across all possible source-target rank configurations, for ranks 1, 2, 4, 8, 16, 32, 64, 128. The overview of the results averaged over the 13 text-only tasks is provided in Figures~\ref{fig:hmap_4b} (Gemma 4B) and Figures~\ref{fig:hmap_1b} (Gemma 1B), with additional results for the VL tasks in Figure~\ref{fig:heatmap_4b_vl}, also in Appendix~\ref{app:additional} (4B). 


The central finding is that fine-tuning a LoRA module at a high source rank ($r_{src}$) and subsequently compressing it to a lower target rank ($r_{tgt}$) on average yields on-par or even higher-quality $r_{tgt}$ LoRA-s (e.g., $16\rightarrow4$ or $128\rightarrow8$ setups) than the ones obtained via direct fine-tuning with $r_{tgt}$; we again remind the reader that we ran hyperparameter sweeps for $r_{tgt}$ to avoid having suboptimal baselines, directly tuned $r_{tgt}$-rank LoRA-s. For the latter two configurations, we observe at least small gains for 9/13 tasks, with substantial gains on some tasks.

\begin{table*}[t!]
\centering
\caption{Performance comparison (Accuracy \%) of different fine-tuning strategies for Gemma 3 4B IT on text tasks. \textit{0-S} refers to zero-shot performance of the base model without any task-specific fine-tuning. +M steps refers to continued fine-tuning after the initial N fine-tuning steps. The best result in each row is highlighted in \textbf{bold}.}
\label{tab:finetuning_strategies_4b_text}
\def\arraystretch{0.999}
{\small
\resizebox{0.999\textwidth}{!}{
\begin{tabular}{@{}lccccccccc@{}}
\toprule
& {} & \multicolumn{3}{c}{\textbf{Direct Fine-tuning ($r=1$)}} & \multicolumn{3}{c}{\textbf{Cont-Squeeze (128 $\rightarrow$ 1)}} & \multicolumn{2}{c}{\textbf{In-Squeeze (128 \ldots $\rightarrow$ 1)}} \\
\cmidrule(lr){3-5} \cmidrule(lr){6-8} \cmidrule(lr){9-10}
\textbf{Task} & \textbf{0-S} & \textbf{+0 steps} & \textbf{+200 steps} & \textbf{+700 steps} & \textbf{+0 steps} & \textbf{+200 steps} & \textbf{+700 steps} & \textbf{Standard} & \textbf{Min steps} \\ \midrule
WNG-L & {\em 27.0} & 90.73 & 90.94 & 90.60 & 90.69 & 91.58 & 91.41 & 91.07 & \textbf{91.88} \\
BOOLQ & {\em 49.8} & 94.39 & 94.28 & 94.34 & 94.11 & 94.39 & 94.06 & 94.30 & \textbf{94.47} \\
PIQA & {\em 35.0} & 93.05 & 93.11 & 93.02 & 92.49 & 92.97 & \textbf{93.22} & 92.89 & 92.80 \\
DROP & {\em 52.6} & \textbf{88.80} & 87.97 & 87.70 & 86.41 & 88.03 & 87.67 & 88.25 & 88.66 \\
ANLI-r2 & {\em 19.9} & 79.30 & 77.23 & 78.57 & 29.30 & 78.46 & 79.52 & 79.63 & \textbf{81.14} \\
PAWS & {\em 35.3} & 97.56 & 97.57 & 96.94 & 74.82 & 97.15 & 97.39 & 97.40 & \textbf{97.60} \\
HSWAG & {\em 50.8} & 97.19 & 97.19 & 97.09 & 96.79 & 97.42 & 97.16 & 97.49 & \textbf{97.51} \\
OBQA & {\em 42.4} & \textbf{94.95} & \textbf{94.95} & 94.83 & 94.71 & 94.47 & 94.71 & 94.59 & 94.71 \\
GoE & {\em 46.0} & 83.25 & 83.54 & 81.91 & 81.32 & 83.26 & 82.73 & 83.83 & \textbf{83.90} \\
ARC-E & {\em 44.8} & 95.12 & \textbf{95.23} & 95.12 & 86.74 & 95.03 & 95.01 & 94.92 & \textbf{95.23} \\
ARC-C & {\em 36.6} & 88.80 & 88.59 & 89.02 & 59.68 & \textbf{89.11} & 88.76 & 88.80 & 88.80 \\
SIQA & {\em 35.8} & 88.78 & 89.35 & 87.87 & 88.13 & 89.33 & 89.35 & 89.77 & \textbf{89.85} \\
MMLU  & {\em 42.6} & 79.19 & 76.61 & 78.69 & 29.97 & 79.43 & \textbf{80.17} & 79.75 & 80.16 \\ \midrule
\textbf{Avg} & {\em 39.9} & 90.08 & 89.74 & 89.67 & 77.32 & 90.05 & 90.09 & 90.21 & \textbf{90.51} \\ \bottomrule
\end{tabular}
}
}%
\end{table*}

\rparagraph{On the Transformation Step and Performance Collapse} The capability of \textit{Post-Squeeze} is a function of (i) its starting rank, where the assumption is that higher ranks would typically provide higher task performance, and (ii) of the actual \textit{transformation step} (i.e., the difference between $r_{tgt}$ and $r_{src}$). The larger the step, the more components get discarded during the RSVD compression, which potentially may result in performance loss or even performance collapse. 
A clear and consistent pattern emerges from the visualizations in Figures~\ref{fig:hmap_4b}-\ref{fig:hmap_1b}. The benefits of P\textit{ost-Squeeze} are most pronounced when compressing from a higher rank to a lower one when the difference between $r_{src}$ and $r_{tgt}$ is (arguably) large but not extreme, and thus not leading to severe information loss. This trend holds true both for 4B and 1B models. While starting from a higher rank is beneficial, the choice of the source rank $r_{src}$ is not arbitrary, and an excessively large gap between $r_{src}$ and $r_{tgt}$ can be detrimental or even lead to a performance collapse (e.g., see the drop for the $128\rightarrow$1 configuration in Figure~\ref{fig:hmap_1b}). This empirical finding has two practical implications:



\noindent First, it directly motivates the \textit{Cont-Squeeze} method, positing the following question: if the collapse is encountered, can performance after the \textit{Post-Squeeze} compression be quickly recovered via short continued fine-tuning? 

\noindent Second, given a desired target rank, there seems to be a `sweet spot' for the source rank, which should be sufficiently large to facilitate good optimization, but not so large that the subsequent compression step becomes overly lossy, and this is also partially model-specific (cf., 4B and 1B variants with $r_{tgt}=2$). We preliminarily analyze the patterns of performance collapse through the lens of variance/energy retention after pruning singular values; see the full discussion in Appendix~\ref{app:variance}. A promising direction for future work is thus to use the rate of retention of the singular values as a proxy to anticipate this collapse and guide the selection of an optimal $r_{src}$ if continued fine-tuning is not possible.

\noindent \textbf{\textit{Cont-Squeeze} and \textit{In-Squeeze} Performance.}
We now compare the performance of continued fine-tuning after \textit{Post-Squeeze} (the \textit{Cont-Squeeze} variant) as well as the effect of gradual in-tuning rank annealing (the general \textit{In-Squeeze} variant). The main experimental setup is as follows: $r_{tgt}=1$, where $r_{src}=128$ for \textit{Cont-Squeeze}, and 128 is also the starting rank for the iterative annealing. For a fair comparison, we keep the total number of training steps equal across all the method variants in the comparison: if the original number of fine-tuning steps for the task was N (e.g., $N=10,000$), for direct fine-tuning we continue fine-tuning for the additional $M=200$ or $M=700$ steps (so that the total number of steps is $N+M$), while for \textit{Cont-Squeeze} we continue fine-tuning also for $M$ steps with $r_{tgt}$ after the $r_{src}\rightarrow r_{tgt}$ transformation. Finally, for the standard \textit{In-Squeeze}, we distribute the $N+M$ steps according to the two schemes described in \S\ref{ss:in_tuning_exp}.

The results with 4B IT on text-only tasks are provided in Table~\ref{tab:finetuning_strategies_4b_text}. The results of the 4B model on the VL tasks and the results with the 1B model are provided in the respective Tables~\ref{tab:finetuning_strategies_4b_vl} and~\ref{tab:finetuning_strategies_1b_text} in Appendix~\ref{app:additional}.  First, we observe that the directly trained rank-1 LoRA seems already saturated; continued fine-tuning offers no performance benefit and can sometimes even lead to minor degradations. In stark contrast, the  `squeezed' (128$\rightarrow$1) LoRA benefits significantly from this brief refinement of \textit{Cont-Squeeze}. The additional fine-tuning steps help it recover from the information loss incurred during the aggressive compression. \textit{Cont-Squeeze} can fully recover performance after mere $M=200$ additional steps, even when the collapse happens (e.g., see the recovery on ANLI-r2 and MMLU). In the case of VL tasks, \textit{Cont-Squeeze} often improves beyond its initial \textit{Post-Squeeze} performance. Overall, this demonstrates that quick continued fine-tuning is an effective and efficient strategy for refining transformed LoRA-s.

Further, the gradual rank annealing, \textit{In-Squeeze}, is proven to be highly effective across the board. As shown in the final two columns of the tables, the `Minimum Steps' sub-variant shows consistently strong performance across all models and task types, and also hits peak performance for the majority of tasks (e.g., 8/13 tasks in Table~\ref{tab:finetuning_strategies_4b_text}). This suggests that a gradual, curriculum-based in-tuning compression schedule is a superior optimization strategy for discovering a robust and high-performing low-rank solution.



\begin{figure}[t!]
    \centering 
        \includegraphics[width=0.99\linewidth]{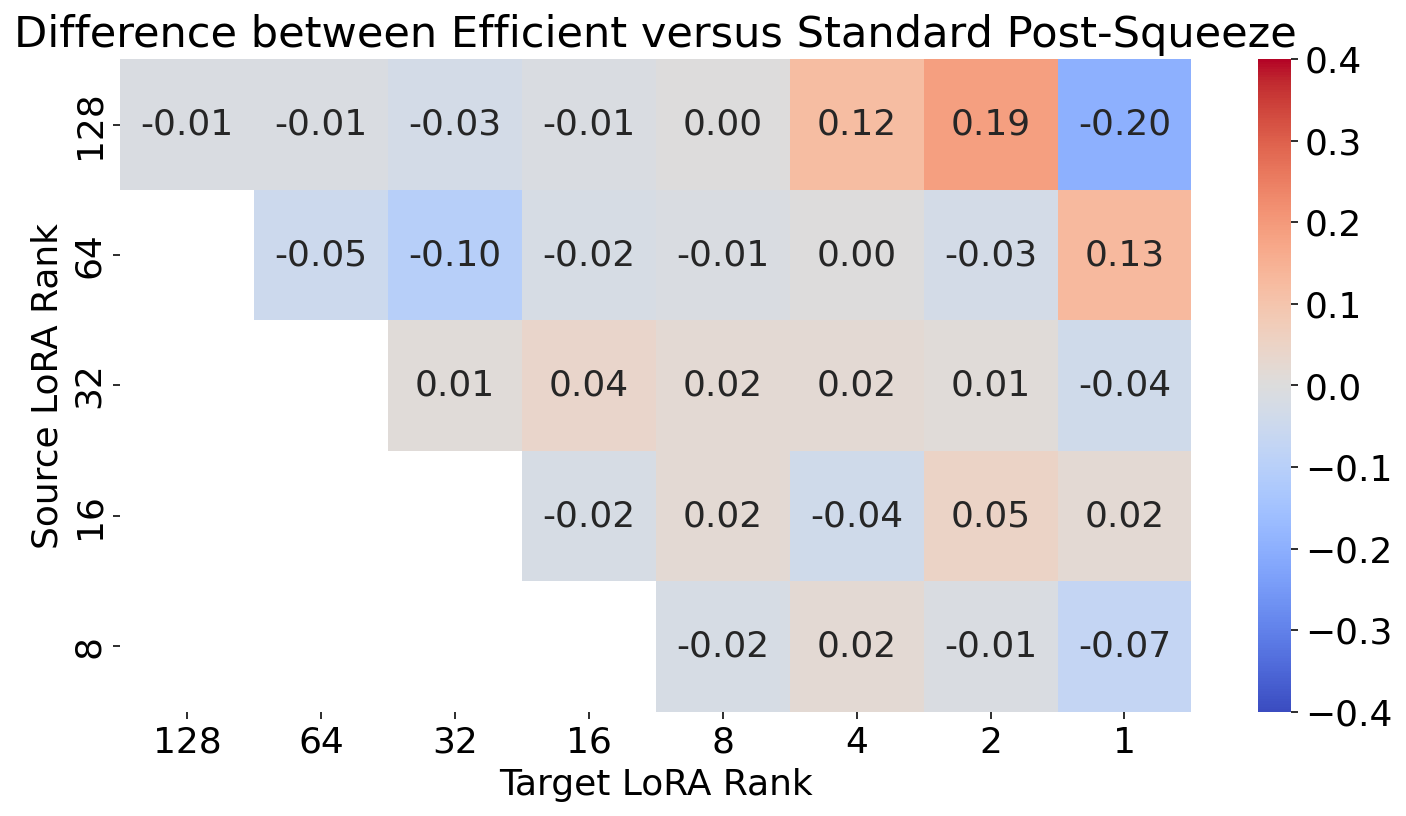}
        \caption{Performance difference between the memory-efficient \method and the standard \textit{Post-Squeeze} variant, averaged over the 13 text-only tasks.}
        \label{fig:efficient_squeeze}
\end{figure}

\subsection{Additional Experiments and Ablations}

\noindent \textbf{Memory-Efficient \method.}
A comparison between memory-efficient \textit{Post-Squeeze} and its standard variant is provided in a heatmap in Figure~\ref{fig:efficient_squeeze}. As a general finding, as expected (see Appendix~\ref{app:derivation}) we do not see any significant difference between the two variants. We further note that the observed patterns (e.g., observed performance collapse or the rate of performance decay) are also very well aligned across individual tasks for the two variants. The tiny fluctuation in the results is only due to non-exactness of Randomized SVD used across the two variants.

\noindent \textbf{(Low) Variation over Random Seeds.}
To ensure that the observed gains with \method (especially with \textit{Cont-Squeeze} and \textit{In-Squeeze}) are not due to the randomness-bound variation of the chosen random seed and RSVD, we also compute the standard error of the mean (SEM) across multiple runs with different random seeds. The SEM scores, as evidenced in Table~\ref{tab:variance_appendix} in Appendix~\ref{app:variance}, are extremely low, and we observe particulary low variance for the lowest ranks. This has two implications: (i) direct fine-tuning with lower ranks is more likely to get stuck in local minima; (ii) absolute gains of \method are typically much larger than the observed SEM scores, confirming that \method can create $r_{tgt}$-rank LoRA-s that avoid those local minima.

\noindent \textbf{What If $r_{tgt}\geq r_{src}$?}
Note that, strictly speaking, it is possible to create a target LoRA where $r_{tgt}\geq r_{src}$. However, the target LoRA cannot capture any new information not captured by the source LoRA. It merely provides a higher-rank SVD-based approximation of the source LoRA unrolled into the full weights space. Basically, the target rank controls the `fine-grainedness' of approximating the approximation of the delta vector provided by the source-rank LoRA.\footnote{The fact that RSVD is the best $r_{tgt}$-rank approximation of the $\Delta W_{src} = A_{src} \times B_{src}$ task vector is also the root cause of slight variance in performance when setting $r_{src}=r_{tgt}$ with \textit{Post-Squeeze}; see the diagonals of Figures~\ref{fig:hmap_4b} and ~\ref{fig:hmap_1b}.} However, this feature of having an arbitrary target rank (including the setting where the target rank is higher than the source rank) might be used, e.g., if a designer wants to have a single LoRA rank across different LoRA-s (trained with different values for $r_{src}$), without the need to retrain them. 

\rparagraph{VL Tasks and Other Model Sizes}
Additional results on text-only tasks (1B) are in Table~\ref{tab:finetuning_strategies_1b_text}, and the results on the VL tasks (4B) are in Table~\ref{tab:finetuning_strategies_4b_vl}; Appendix~\ref{app:additional}. The key findings, with some slight differences (e.g., the performance on the VL tasks seems more saturated and the gains with \method, while still visible are less pronounced than on text-only tasks), still hold. As expected, the performance collapse with too large transformation steps (e.g., $128\rightarrow 1$) is more salient with a smaller, 1B model - nonetheless, \textit{Cont-Squeeze} can again recover performance extremely quickly after 200 additional fine-tuning steps and \textit{In-Squeeze} remains the most powerful model variant on average (see Table~\ref{tab:finetuning_strategies_1b_text}). Finally, a small experiment with \textit{Post-Squeeze}) on 12B for a selection of tasks further confirms the robustness of \method; see Figure~\ref{fig:12b} in Appendix~\ref{app:additional}.

Finally, we also provide a brief discussion on theoretical implications (e.g., relationship to lottery ticket hypothesis) of \method in Appendix~\ref{app:theoretical}.





%% file: sections/06_conclusion_google.tex
\section{Conclusion}
We introduced {\em \method}, a simple yet effective methodology for compressing LoRA modules either post-hoc or dynamically during fine-tuning. Our experiments over a suite of 13 text-based and 10 vision-language VQA tasks demonstrated that it is often more advantageous to fine-tune with a higher, `overparameterized' LoRA rank and then compress to a lower target rank, rather than fine-tuning at the fixed target rank directly. This approach not only yields modules with a better size-to-performance trade-off but also simplifies deployment by decoupling the training rank from the deployment rank and reducing the need for rank-specific hyperparameter tuning. The gradual, in-tuning rank annealing variant delivered the most robust results in general.


Due to its wide adoption, extensive use and architectural simplicity, in this work we focused on the standard LoRA design. For future work, one research direction is to extend the LoRA-Squeeze principles beyond standard LoRA setups to more sophisticated LoRA variants. We also plan to utilize \method principles in techniques for module merging~\citep{stoica2025iclr}. Furthermore, we plan to explore methods for automatically determining the optimal source rank or gradual rank annealing schedules for a given target rank by extending our preliminary analyses (see Figure~\ref{fig:ret_tasks} in the Appendix) on the rate of decay of singular values during decomposition. 

%% file: sections/xx_appendix_icml.tex
\section{Learning Rate Selection Details}
\label{app:lr_sweep}

For all direct fine-tuning experiments on text tasks, we first determined the optimal learning rate for each LoRA rank and model size combination to ensure that our baseline comparisons were as strong as possible. We used the Adafactor optimizer with a linear warmup of 1000 steps and no subsequent learning rate decay.

The optimal learning rate was selected via a grid search over the set \{0.001, 0.003, 0.01, 0.03, 0.1\}. For each (model, rank) pair, we chose the learning rate that yielded the highest average performance across all 12 of our text-only tasks. This process revealed a consistent trend: lower-rank LoRA configurations (e.g., $r \le 4$) benefited from a higher learning rate, whereas higher-rank configurations achieved better performance with a more conservative rate.

The specific learning rates identified and used for the baseline models in our experiments are detailed in Table~\ref{tab:lr_sweep_appendix}.

\begin{table}[h!]
\centering
\caption{Optimal learning rates determined by grid search for different model sizes and LoRA ranks on the text task suite. The chosen rate maximized the average performance across all tasks.}
\label{tab:lr_sweep_appendix}
\begin{tabular}{@{}lcc@{}}
\toprule
\textbf{Model Size} & \textbf{LoRA Rank ($r$)} & \textbf{Optimal LR} \\ \midrule
\multirow{2}{*}{Gemma 3 4B} & 1, 2, 4 & 0.03 \\
 & $\ge 8$ & 0.01 \\ \midrule
\multirow{2}{*}{Gemma 3 1B} & 1, 2 & 0.03 \\
& $\ge 8$ & 0.01 \\ \midrule
\multirow{1}{*}{Gemma 3 4B VL} & $\ge 1$ & 0.01 \\
\bottomrule
\end{tabular}
\end{table}

\section{Varianc over Random Seeds}
\label{app:variance}
The results in Table~\ref{tab:variance_appendix} are obtained over 3 random seeds: \{42, 43, 44\}.

\begin{table}[h!]
\centering
\caption{Average performance and standard error of the mean (SEM) across all text tasks for different LoRA ranks and learning rates for Gemma 3 4B.}
\label{tab:variance_appendix}
\begin{tabular}{@{}cccc@{}}
\toprule
\textbf{LoRA Rank ($r$)} & \textbf{LR} & \textbf{Acc (\%)} & \textbf{SEM} \\ \midrule
\multirow{2}{*}{1} & 0.01 & 89.96 & 0.061 \\
 & 0.03 & 90.21 & 0.061 \\ \midrule
\multirow{2}{*}{32} & 0.01 & 90.55 & 0.075 \\
 & 0.03 & 88.02 & 0.080 \\ \midrule
\multirow{2}{*}{128} & 0.01 & 90.67 & 0.010 \\
 & 0.03 & 80.87 & 0.127 \\ \bottomrule
\end{tabular}
\end{table}




\section{Evaluation Tasks and Datasets}
\label{app:tasks}
The summary of text-based and VL tasks is provided in Tables~\ref{tab:datasets} and~\ref{tab:vl-datasets}.

\begin{table*}[t!]
\centering
\caption{Summary of text-based evaluation datasets.}
\label{tab:datasets}
{\small
\begin{tabular}{lllll}
\toprule
\textbf{Dataset} & \textbf{Abbreviation} & \textbf{Reference} & \textbf{Train Split} & \textbf{Dev /Test Split} \\
\midrule
WinoGrande & WNG-L* & \citet{sakaguchi2020winogrande} & 10,234* & 1,267 / 1,767 \\
BoolQ & BOOLQ & \citet{clark-etal-2019-boolq} & 9,427 & 3,270 / 3,245 \\
PIQA & PIQA & \citet{bisk2020piqa} & 16,113 & 1,838 / 3,084 \\
DROP & DROP & \citet{duan-etal-2019-drop} & 77,409 & 9,536 / 9,653** \\
ANLI (Round 2) & ANLI-r2 & \citet{nie-etal-2020-adversarial} & 45,460 & 1,000 / 1,000 \\
PAWS Wiki & PAWS & \citet{zhang-etal-2019-paws} & 49,401 & 8,000 / 8,000 \\
HellaSwag & HSWAG & \citet{zellers-etal-2019-hellaswag} & 39,905 & 10,004 / 10,042 \\
OpenBookQA & OBQA & \citet{mihaylov-etal-2018-suit} & 4,957 & 500 / 500 \\
GoEmotions & GoE & \citet{demszky-etal-2020-goemotions} & 43,410 & 5,426 / 5,427 \\
ARC-Easy & ARC-E & \citet{clark2018think} & 2,251 & 570 / 2,376 \\
ARC-Challenge & ARC-C & \citet{clark2018think} & 1,119 & 299 / 1,172 \\
SocialIQA & SIQA & \citet{sap-etal-2019-social} & 33,410 & 1,954 / 2,223 \\
MMLU & MMLU & \citet{hendrycks2021measuring} & 99,842*** & 1,540 / 14,042 \\
\bottomrule
\end{tabular}%
}%
\begin{flushleft}
\small
* We use the \textit{train\_l} (large) subset for fine-tuning. \\
** The DROP test set is hidden and requires submission to a leaderboard for evaluation, we thus evaluate on the development set \\
*** MMLU is an evaluation benchmark and does not have a designated training set; we fine-tune LoRA-s on the standard \textit{auxiliary} training set.
\end{flushleft}
\end{table*}

\begin{table*}[t!]
\centering
\caption{Summary of VL evaluation datasets.}
\label{tab:vl-datasets}
{\small
\begin{tabular}{llll}
\toprule
\textbf{Dataset} & \textbf{Reference} & \textbf{Train Split} & \textbf{Dev /Test Split} \\
\midrule
AI2D & \citet{kembhavi2016ai2d} & 8,129 & 1,000 / 1,000 \\
A-OKVQA & \citet{schwenk2022aokvqa} & 17,056 & 1,146 / 4,524 \\
CountBenchQA & \citet{gemma2024paligemma} & {} & {} \\
DocVQA & \citet{mathew2021docvqa} & 39,463 & 5,349 / 5,188 \\
InfoVQA & \citet{mathew2022infographicvqa} & 31,481 & 4,342 / 8,690 \\
OCR-VQA & \citet{mishra2019ocr} & 801,316 & 95,572 / 101,632 \\
OK-VQA & \citet{marino2019okvqa} & 9,009 & 5,046 / N/A* \\
ScienceQA & \citet{lu2022learn} & 12,726 & 4,241 / 4,241 \\
TallyQA & \citet{acharya2019tallyqa} & 252,477 & N/A** / 53,941 \\
TextVQA & \citet{singh2019textvqa} & 34,602 & 5,000 / 7,764 \\
\bottomrule
\end{tabular}
}%
\begin{flushleft}
\small
* OK-VQA has no public test split; evaluation is performed on the validation split. \\
** TallyQA does not have a public validation split.
\end{flushleft}
\vskip -0.1in
\end{table*}

\section{Derivation of Memory-Efficient \method}
\label{app:derivation}

Here, we provide further details on the derivation of the memory-efficient variant of \method, described in Algorithm~\ref{alg:efficient_squeeze} in the main paper. We seek to compute SVD of $\Delta W_{src}$ (i.e., the full model update as approximated by the source-rank LoRA) to truncate the least-contributing dimensions efficiently without explicitly constructing the (typically large) dense $m \times n$ matrix.




For simplicity of notation, in the following derivation we assume that $r=r_{src}$, $\Delta W = \Delta W_{src}$, $A=A_{src}$, and $B = B_{src}$. We first perform QR decomposition on the factor matrices $A$ and $B$ to isolate their orthonormal bases:
\begin{align}
    & A = Q_A R_A \quad \text{where } Q_A \in \mathbb{R}^{m \times r}, R_A \in \mathbb{R}^{r \times r} \\
    & B^\top = Q_B R_B \quad \text{where } Q_B \in \mathbb{R}^{n \times r}, R_B \in \mathbb{R}^{r \times r}
\end{align}
By definition of the QR decomposition, $Q_A$ and $Q_B$ have orthonormal columns:
\begin{equation}
    Q_A^\top Q_A = I_r, \quad Q_B^\top Q_B = I_r
\end{equation}
We then substitute the decompositions back into the expression for $\Delta W$:
\begin{align}
    \Delta W &= (Q_A R_A) (R_B^\top Q_B^\top) \\
    \Delta W &= Q_A (R_A R_B^\top) Q_B^\top, 
\end{align}
and we can then define the core interaction matrix $M \in \mathbb{R}^{r \times r}$ as:
\begin{equation}
    M = R_A R_B^\top
\end{equation}
Thus, $\Delta W = Q_A M Q_B^\top$.

We then perform (full or randomized) SVD on the small matrix $M^{r\times r}$:
\begin{equation}
    M = U_M S_M V_M^\top
\end{equation}
where $U_M, V_M \in \mathbb{R}^{r \times r}$ are orthogonal matrices and $S_M$ is the diagonal matrix of singular values.

Substituting this back into $\Delta W$:
\begin{equation}
    \Delta W = Q_A (U_M S_M V_M^\top) Q_B^\top
\end{equation}
If we regroup the factor matrices above, we can write:
\begin{equation}
\Delta W = \underbrace{(Q_A U_M)}_{\text{New } U} S_M \underbrace{(V_M^T Q_B^T)}_{\text{New } V^T}
\label{eq:final}
\end{equation}

Since $Q_A$ is orthogonal and $U_M$ is orthogonal, their product $(Q_A U_M)$ is also orthogonal. The same applies to the right side. Therefore, $S_M$ is exactly equal to $S$ derived from directly applying SVD on $\Delta W$. 

If we then prune the rank of matrices in Eq.~\eqref{eq:final} to $r_{tgt}$, we then create an $r_{tgt}$-rank approximation of $\Delta W_{src}$, which is exactly the central mechanism of \method.

\section{On (Coarsely Approximated) Computational Complexity of Different Decompositions}
\label{app:complexity}
Doing 1) full SVD decomposition on matrix $\Delta W^{m \times n}$ has a total complexity of $O(mn~\text{min}(m,n))$, while 2) using Randomized SVD instead decreases it to $O(mn~(r_{tgt}+k_o))$. Finally, 3) the efficient variant from Algorithm~\ref{alg:efficient_squeeze} decreases it further to $O((m+n)~r_{src}^2)$. Let's assume that $m=2,048, n=2,048, r_{src}=64, r_{tgt}=8, k_o = 10$. In this case, the total, theoretically estimated number of FLOPs required for version 1 (full SVD on $\Delta W$) is $\sim$$8.6B$ FLOPs, it is $\sim$$75.5M$ FLOPs for version 2 (randomized SVD on $\Delta W$), and it is only $\sim$$16.8M$ FLOPs for the memory-efficient \method variant.\footnote{Note that versions 1 and 2 also need to spend additional $m \times n \times r_{src}$ FLOPs on creating $\Delta W$, which is bypassed by the memory-efficient variant.}

\section{On Task Knowledge Retention}
We hypothesise that the collapse observed with some tasks when the \textit{Post-Squeeze} transformation step is too large (e.g., 128$\rightarrow$1, see Figures~\ref{fig:hmap_4b} and~\ref{fig:hmap_1b} in the main paper) is due to discarding too much information. We quantify the amount of kept task-related information from $r_{src}$ to $r_{tgt}$ as \textit{variance/energy retention}, and compute it based on sorted singular values:
\begin{align}
V_r(r_{src}\rightarrow r_{tgt}) = \frac{\sum_{i=1}^{r_{tgt}} s_i^2}{\sum_{i=1}^{r_{src}} s_i^2},
\end{align}
where $s_i$ denotes the $i$-th singular value from matrix $S$ obtained via RSVD. By design, $V_r = 100\%$ when $r_{src} = r_{tgt}$ and then it starts decreasing with the decrease of $r_{tgt}$.

\begin{figure}[h]
\vskip 0.2in
    \centering
    \includegraphics[width=0.48\columnwidth]{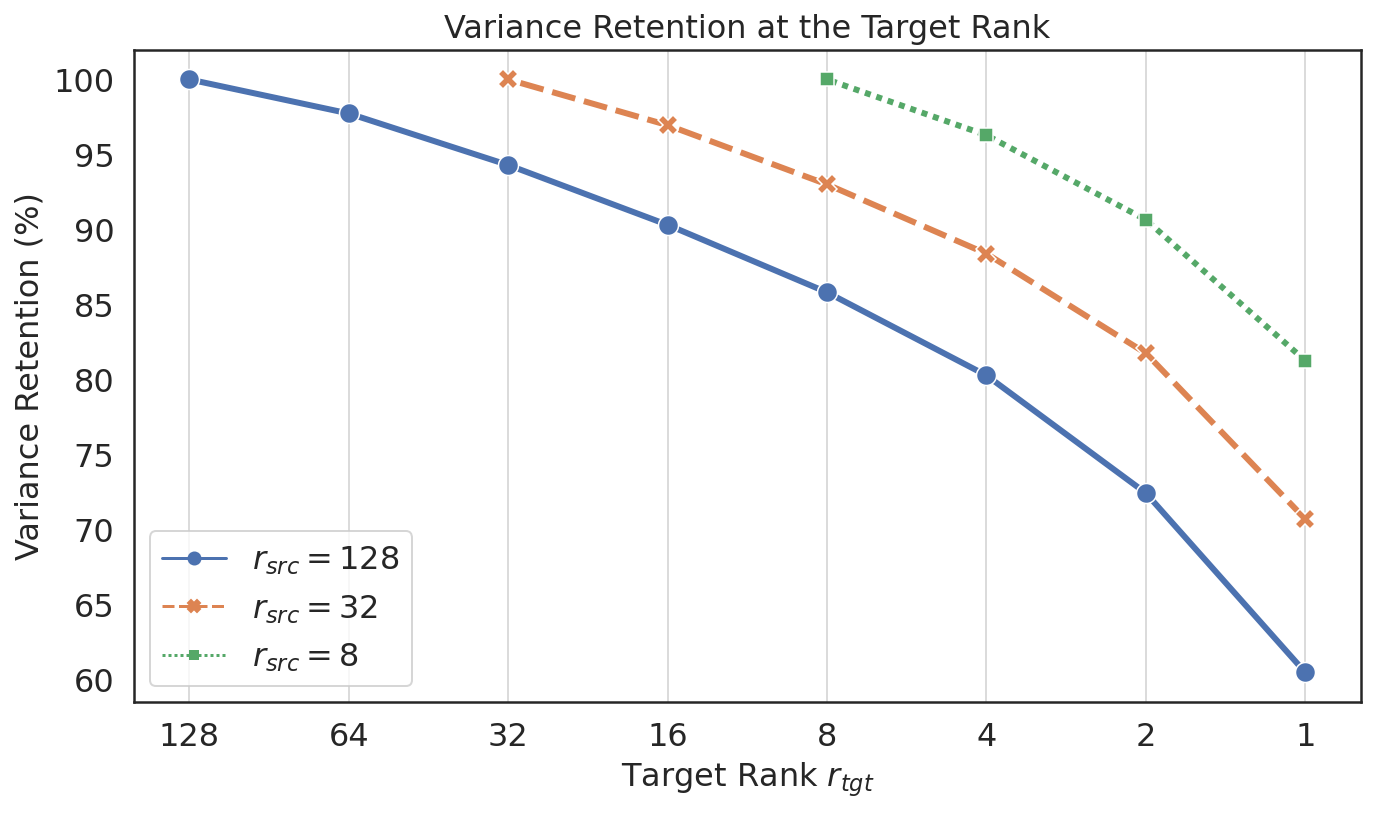}
    \caption{Variance retention scores averaged over 13 text-only tasks for 3 different source ranks $r_{src}$. Per-task scores are provided in Figures~\ref{fig:ret_ar2}-\ref{fig:ret_wng_l} in the appendix.}
    \label{fig:avg_retention}
\end{figure}

\begin{figure*}[h!] 
    \centering 
        \begin{subfigure}[h]{0.32\textwidth}
        \centering
        \includegraphics[width=0.99\linewidth]{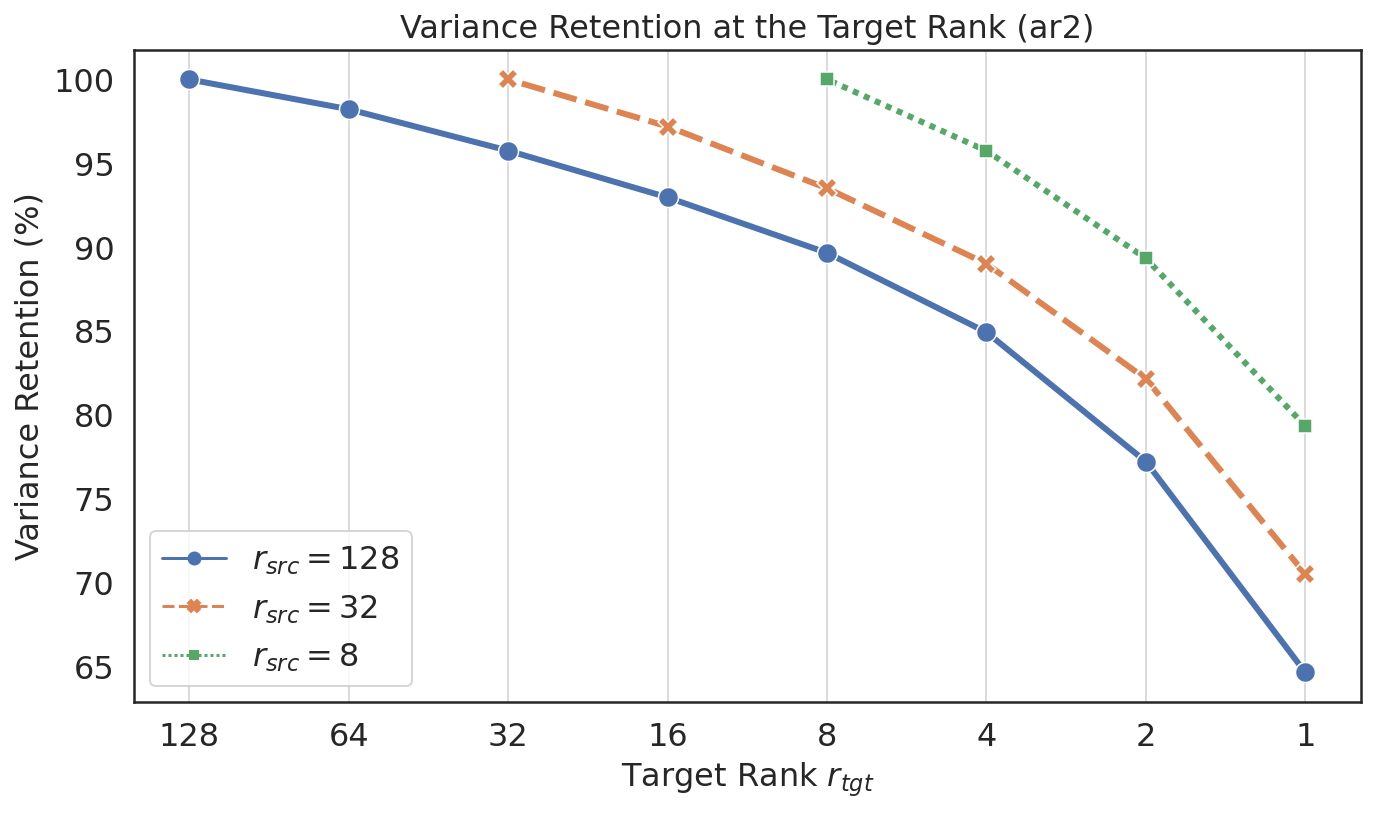}
        \caption{ANLI-r2}
        \label{fig:ret_ar2}
    \end{subfigure}
    \hfill 
    \begin{subfigure}[h]{0.32\textwidth}
        \centering
        \includegraphics[width=0.99\linewidth]{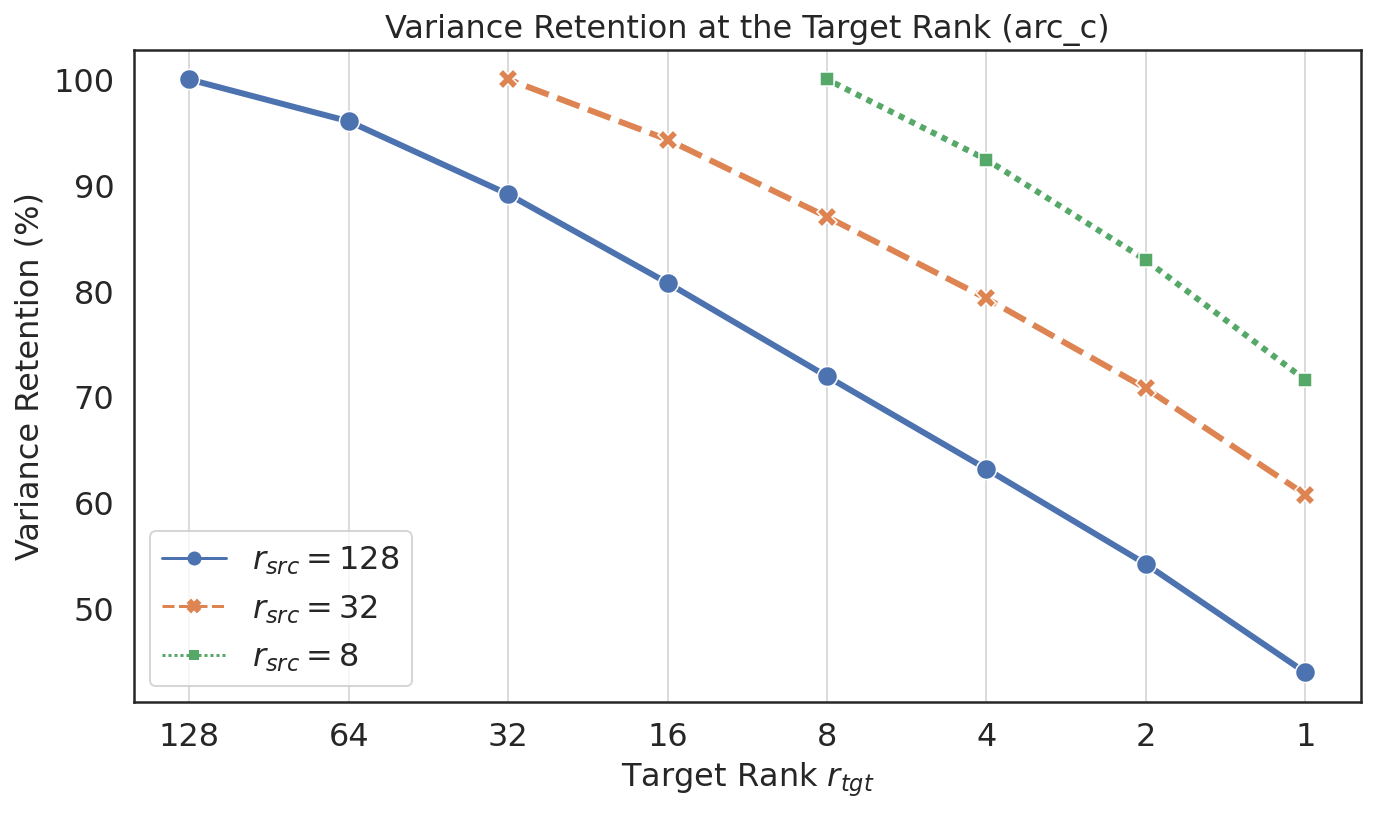}
        \caption{ARC-C}
        \label{fig:ret_arc_c}
    \end{subfigure}
    \hfill
    \begin{subfigure}[h]{0.32\textwidth}
        \centering
        \includegraphics[width=0.99\linewidth]{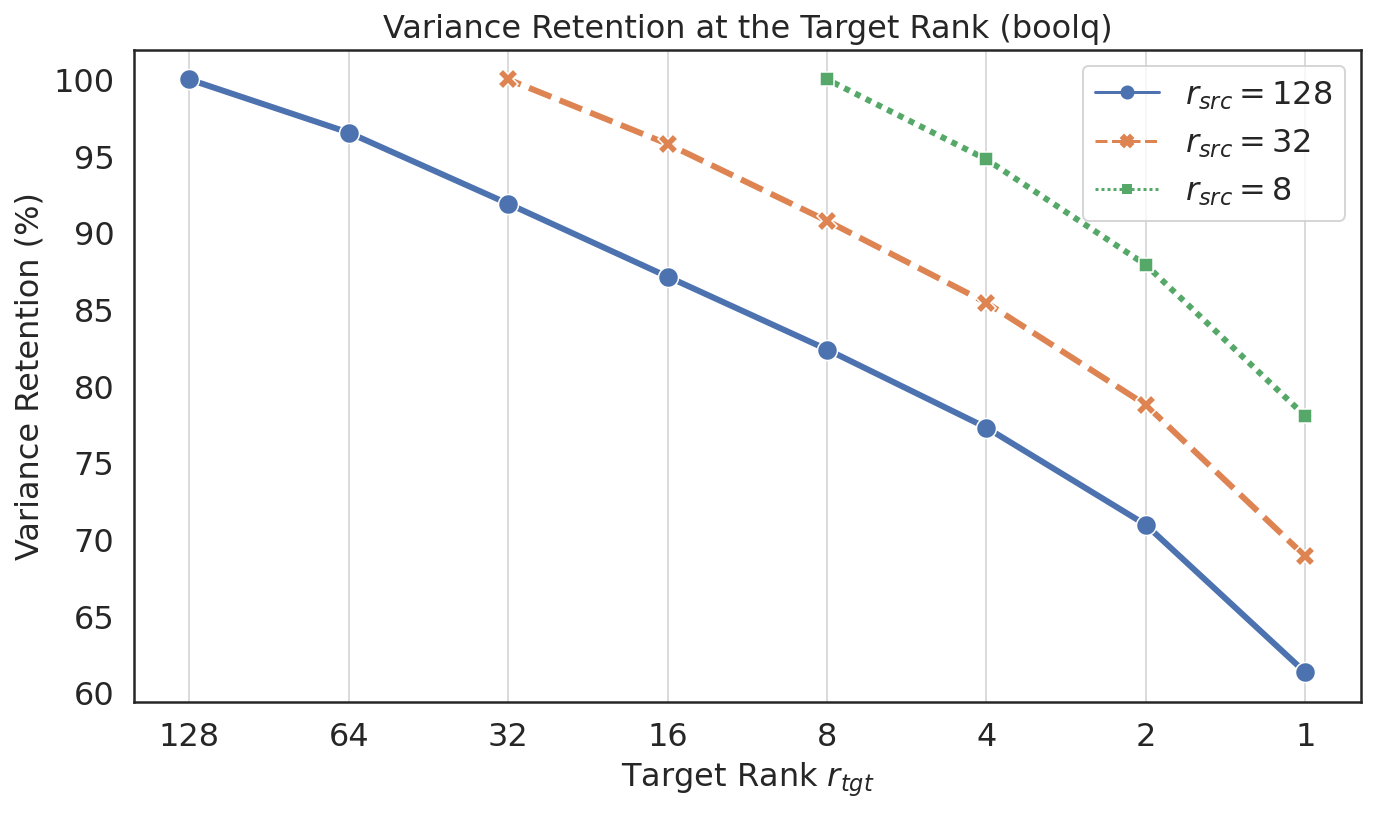}
        \caption{BoolQ}
        \label{fig:ret_boolq}
    \end{subfigure}
    \hfill
            \begin{subfigure}[h]{0.32\textwidth}
        \centering
        \includegraphics[width=0.99\linewidth]{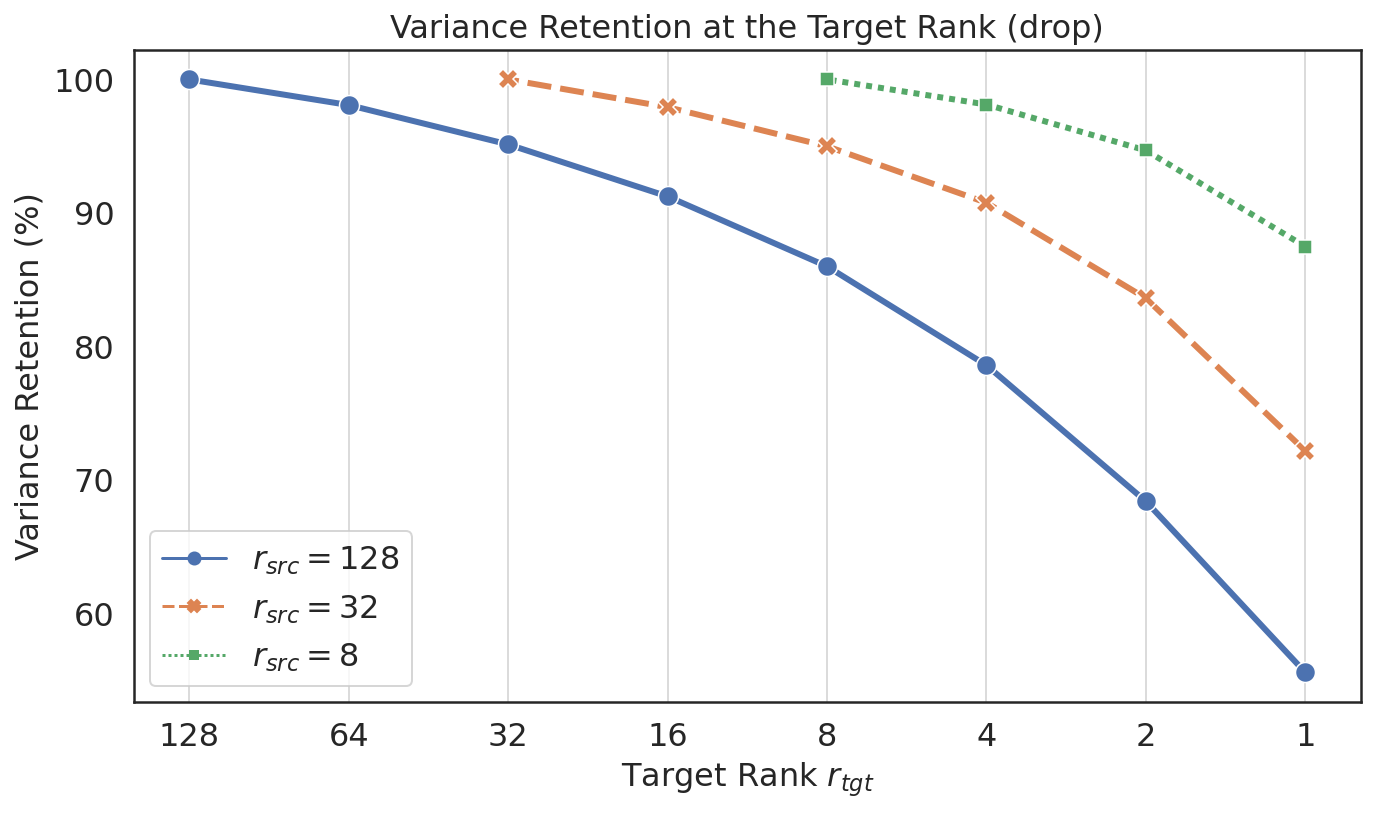}
        \caption{DROP}
        \label{fig:ret_drop}
    \end{subfigure}
    \hfill 
    \begin{subfigure}[h]{0.32\textwidth}
        \centering
        \includegraphics[width=0.99\linewidth]{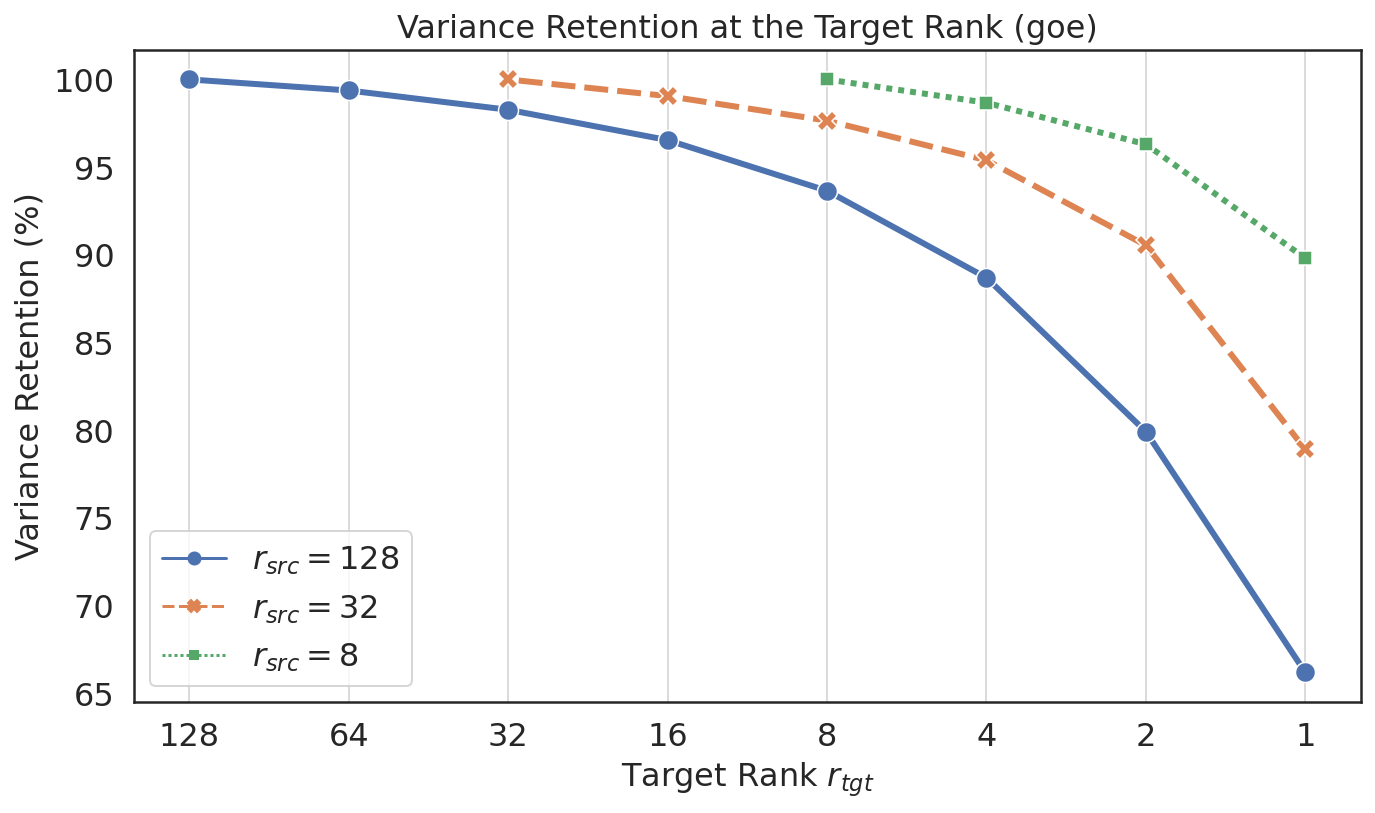}
        \caption{GoE}
        \label{fig:ret_ge}
    \end{subfigure}
    \hfill
    \begin{subfigure}[h]{0.32\textwidth}
        \centering
        \includegraphics[width=0.99\linewidth]{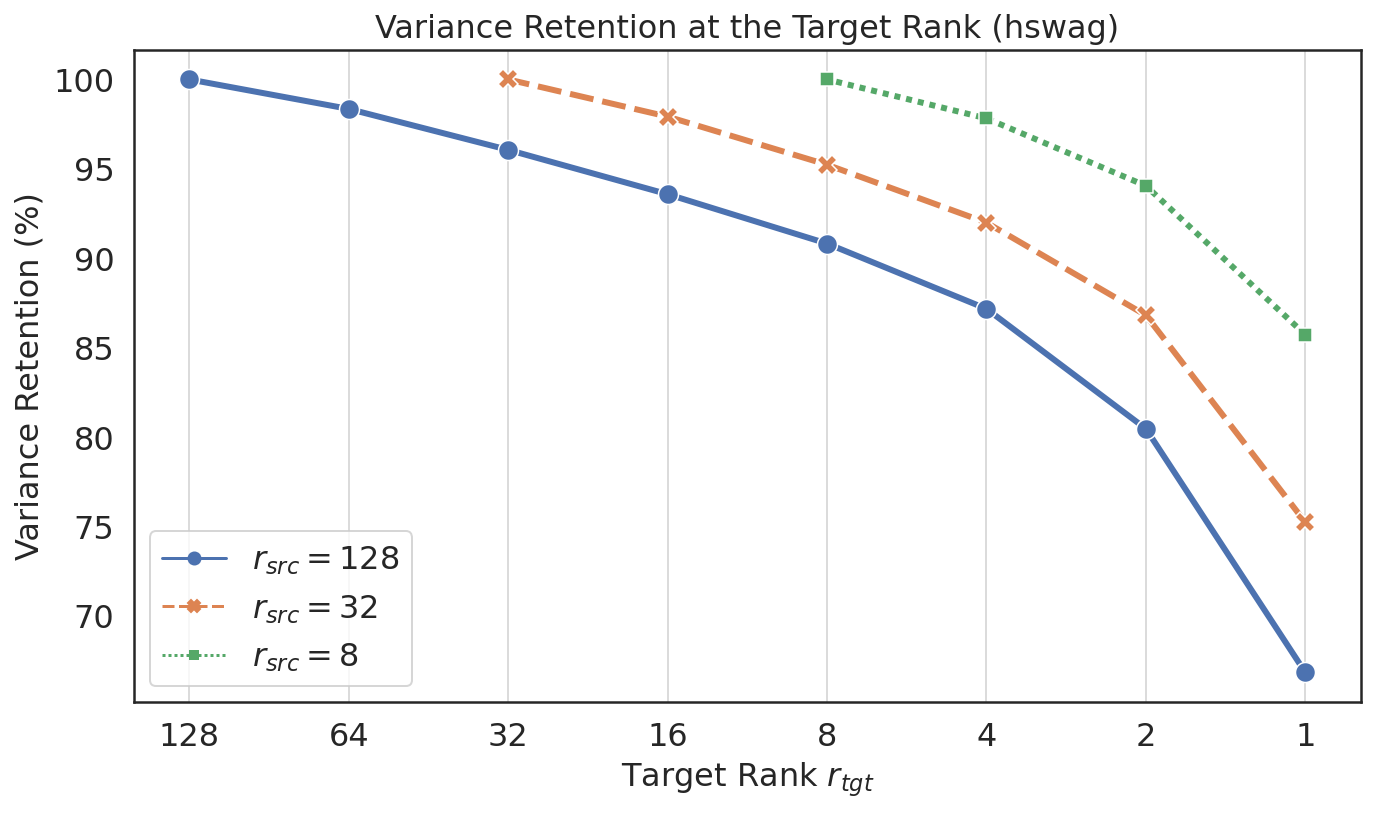}
        \caption{HSWAG}
        \label{fig:ret_hswag}
    \end{subfigure}
    \hfill
            \begin{subfigure}[h]{0.32\textwidth}
        \centering
        \includegraphics[width=0.99\linewidth]{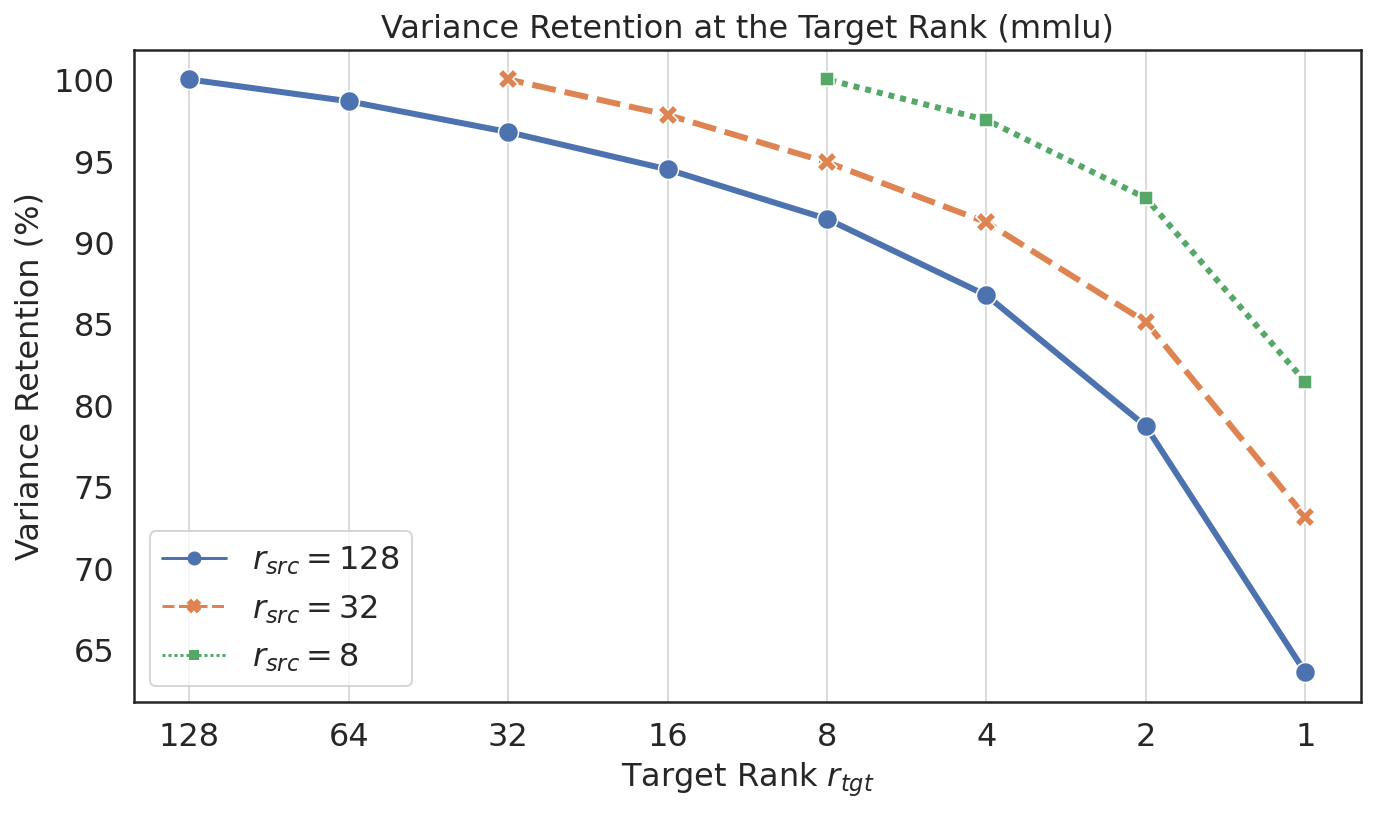}
        \caption{MMLU}
        \label{fig:ret_mmlu}
    \end{subfigure}
    \hfill 
    \begin{subfigure}[h]{0.32\textwidth}
        \centering
        \includegraphics[width=0.99\linewidth]{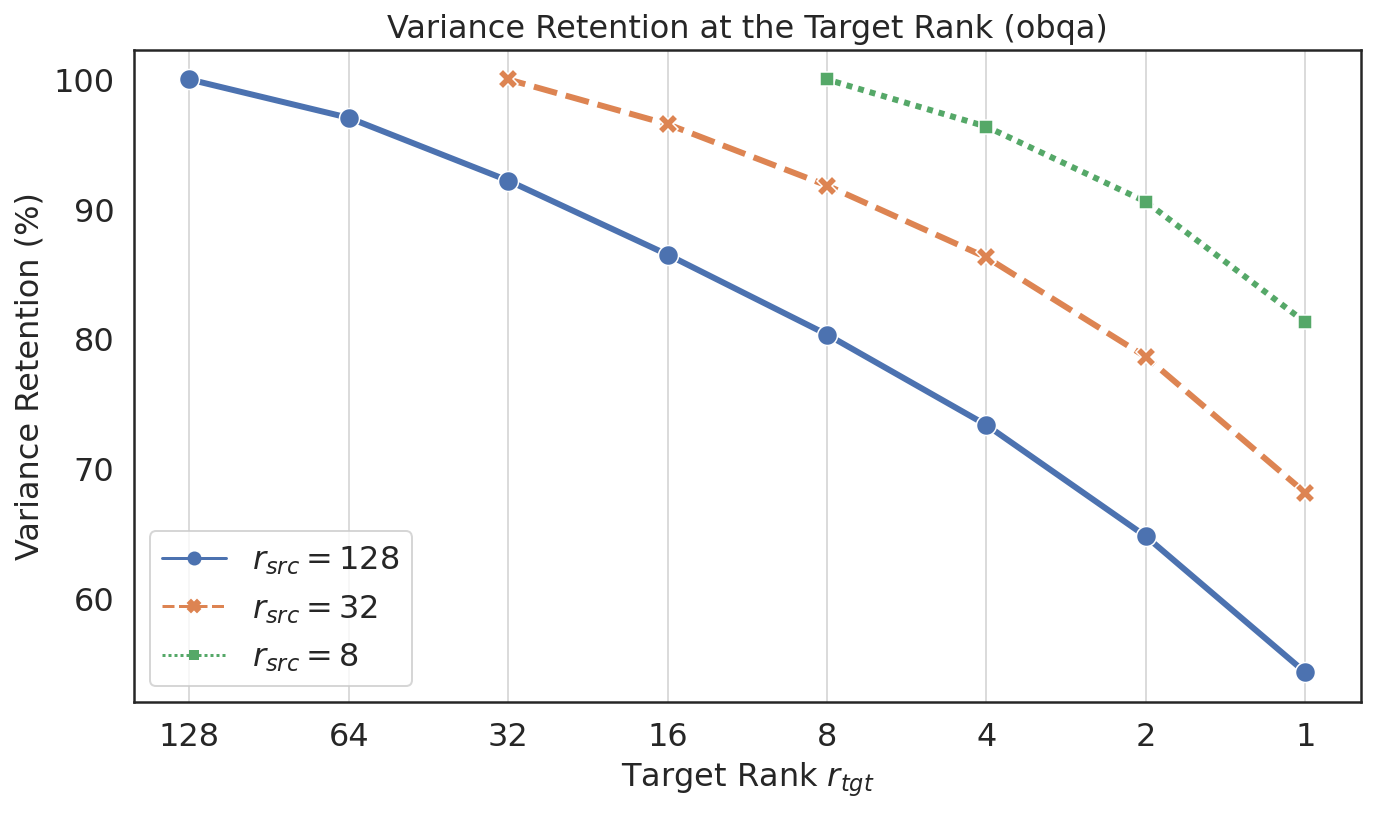}
        \caption{OBQA}
        \label{fig:ret_obqa}
    \end{subfigure}
    \hfill
    \begin{subfigure}[h]{0.32\textwidth}
        \centering
        \includegraphics[width=0.99\linewidth]{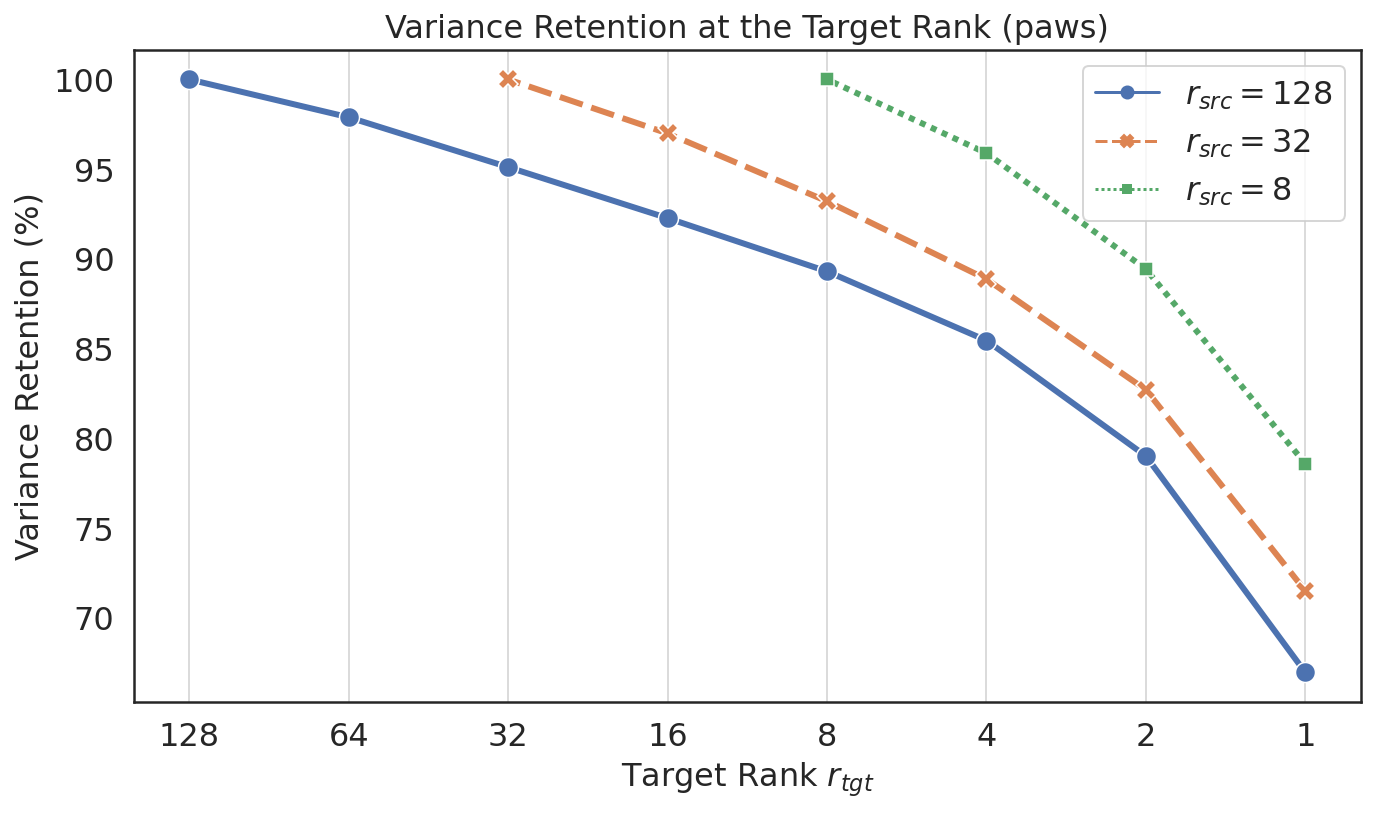}
        \caption{PAWS}
        \label{fig:ret_paws}
    \end{subfigure}
    \hfill
            \begin{subfigure}[h]{0.32\textwidth}
        \centering
        \includegraphics[width=0.99\linewidth]{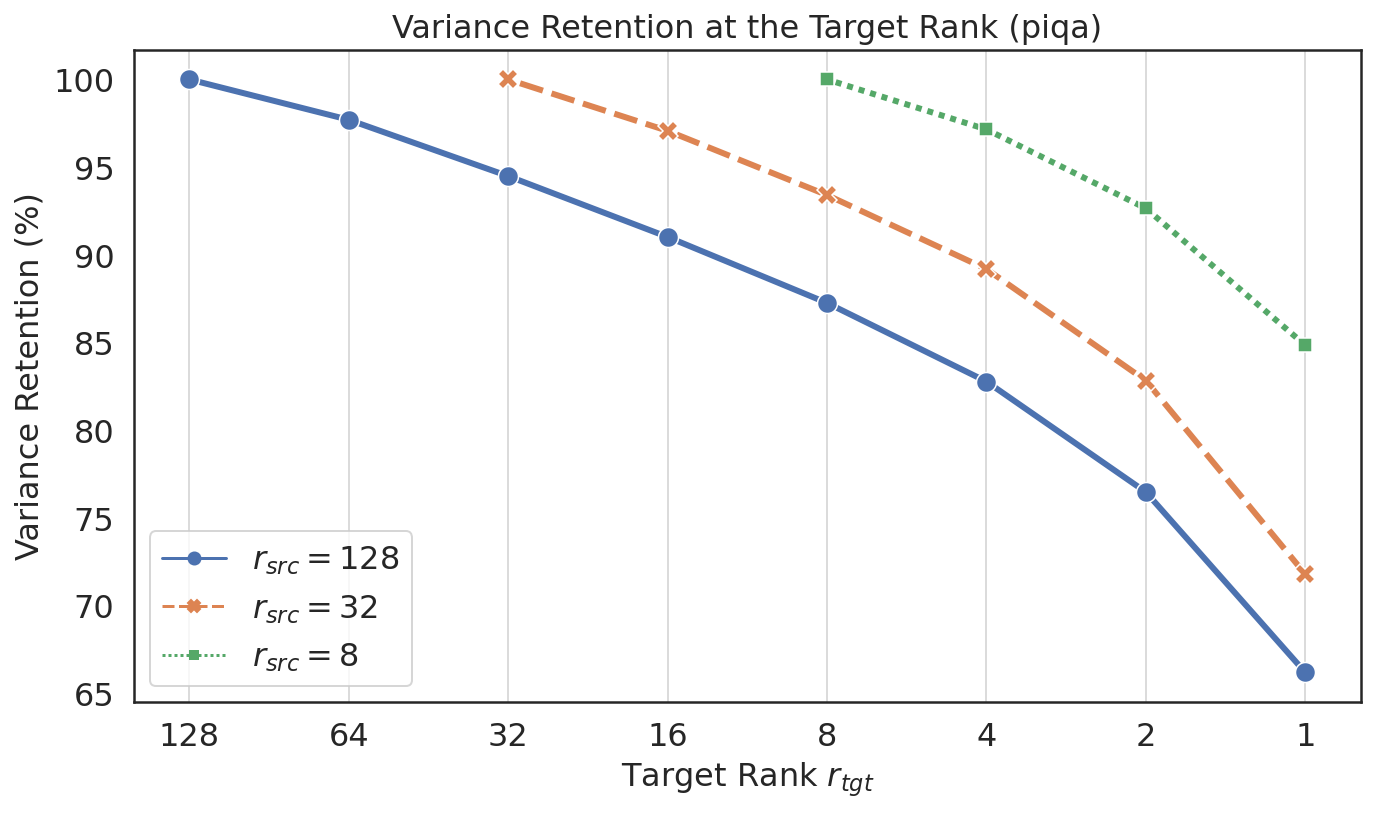}
        \caption{PIQA}
        \label{fig:ret_piqa}
    \end{subfigure}
    \hfill 
    \begin{subfigure}[h]{0.32\textwidth}
        \centering
        \includegraphics[width=0.99\linewidth]{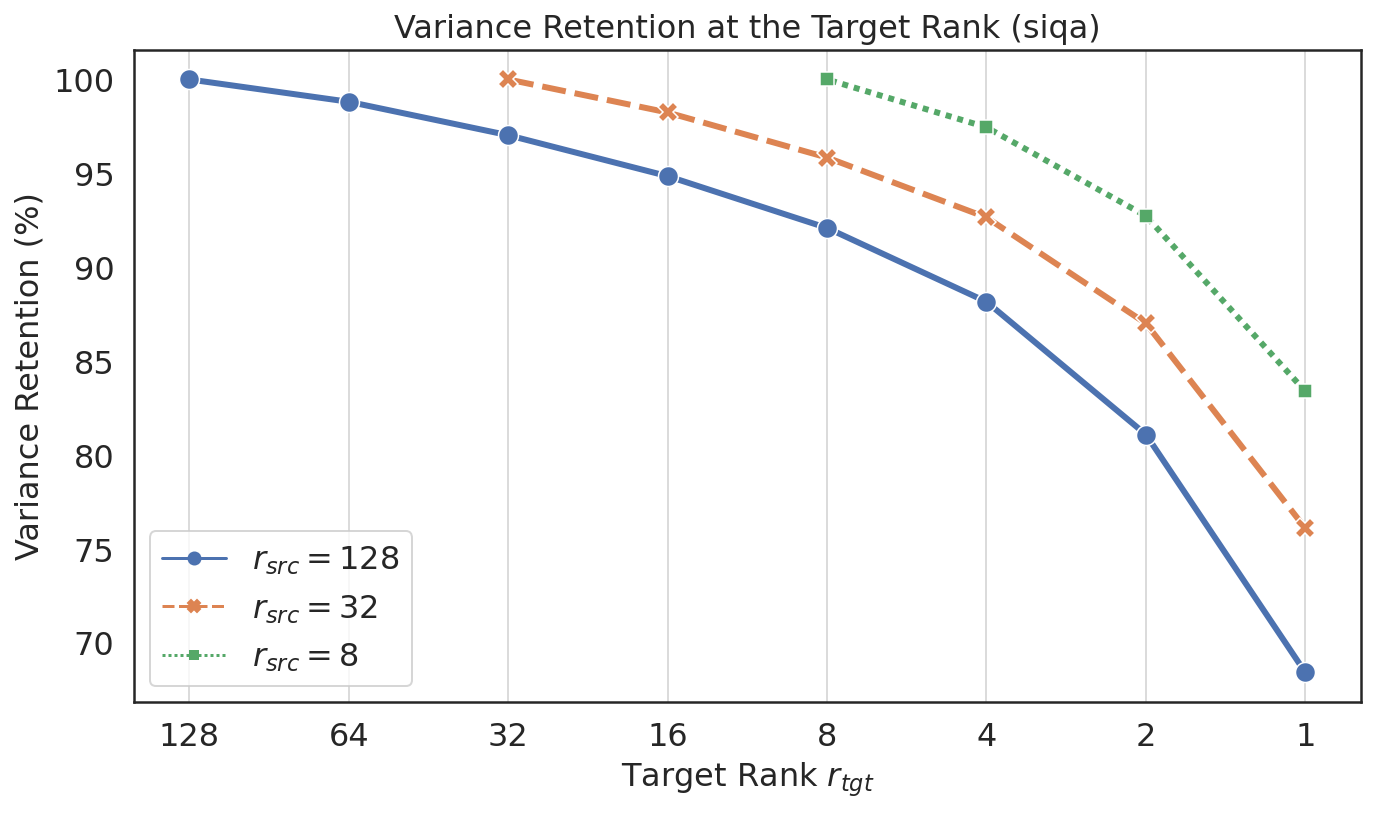}
        \caption{SIQA}
        \label{fig:ret_siqa}
    \end{subfigure}
    \hfill
    \begin{subfigure}[h]{0.32\textwidth}
        \centering
        \includegraphics[width=0.99\linewidth]{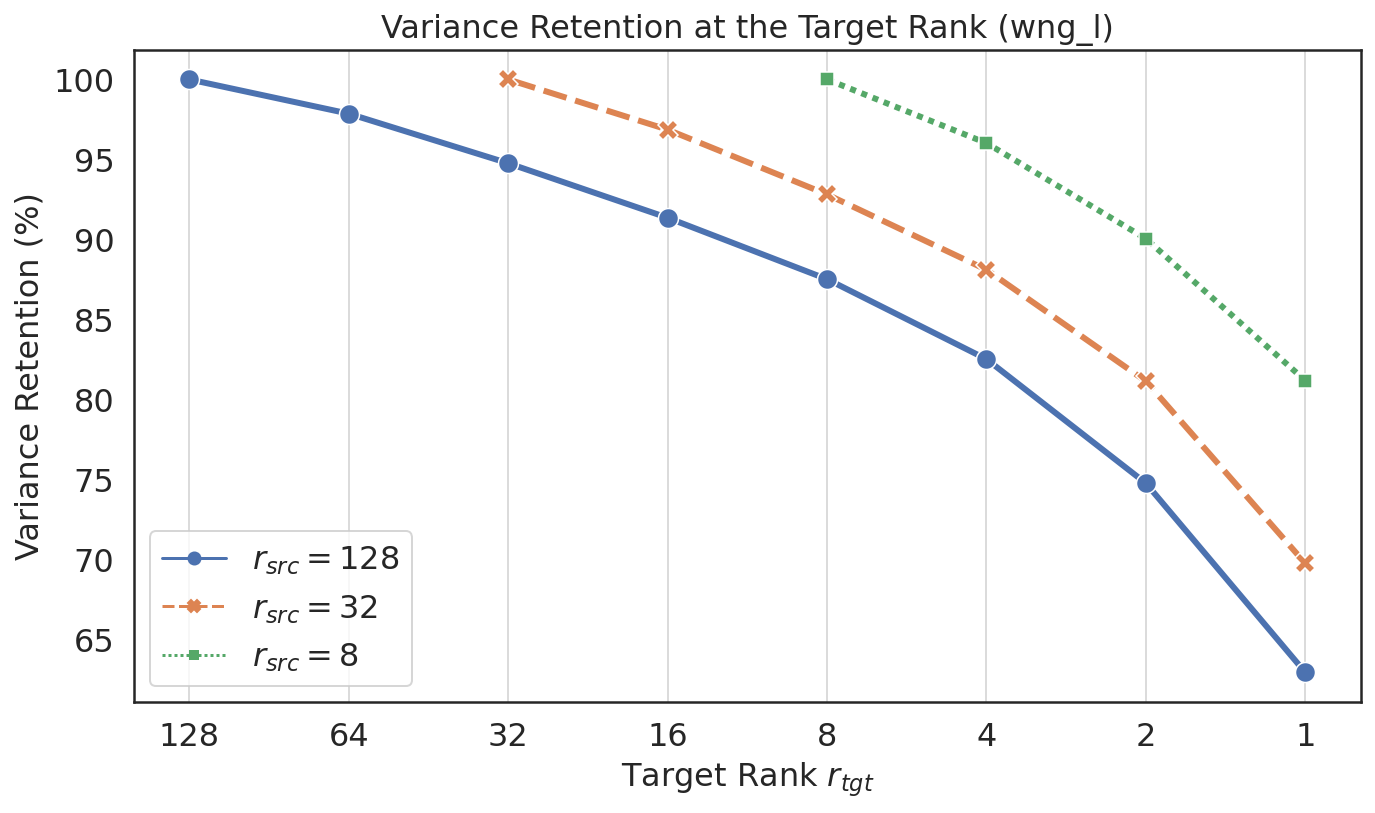}
        \caption{WNG-L}
        \label{fig:ret_wng_l}
    \end{subfigure}

    \caption{Per-task variance retention scores for 12 text-only tasks for 3 different source ranks $r_{src}$. The scores for ARC-E are not shown for clarity of presentation and because they closely align with the shown curves for ARC-C.}
    \label{fig:ret_tasks}
\end{figure*}

Variance retention scores averaged across the 13 text-only tasks for 3 different $r_{src}$ ranks are provided in Figure~\ref{fig:avg_retention} in while per-task $V_r$ are available in Figures~\ref{fig:ret_ar2}-\ref{fig:ret_wng_l}.

In general, we see unequal retention curves for different tasks, and the tasks for which the retention is the lowest for large transformation steps yield the observed performance collapse (e.g., ARC-C, MMLU). On the other hand, there is a large retention for smaller transformation steps: e.g., $V_r$ scores when halving the rank are $\sim96\%+$ ($r_{src}=128$) or $\sim94\%+$ ($r_{src}=32$) for all the tasks, and retention, although it depends on $r_{src}$, remains high when $r_{tgt} = r_{src}/4$ and $r_{tgt} = r_{src}/8$, which explains why there is no performance collapse for transformation steps of this magnitude. In future research, we also hope to further analyze variance retention as a proxy towards quantifying task complexity.

\section{Further Discussion: On Theoretical Motivation and Connections}
\label{app:theoretical}

\noindent \textbf{Connections to Intrinsic Task Dimensionality.} As established in prior work, the main motivation for the LoRA-style task adaptation relies on the fact that the adaptation lies on a low-dimensional manifold within the high-dimensional parameter space~\citep{aghajanyan2021intrinsic, ansuini2019intrinsic}. This implies that the task vector $\Delta W$ has a low intrinsic rank. The objective of LoRA training is to find a good low-rank approximation of this vector. However, directly optimizing within a low-rank space (i.e., training LoRA with a small, fixed rank $r_{tgt}$ from the start) is a non-convex and highly constrained optimization problem. Such a process can easily converge to a suboptimal local minimum, failing to fully capture the essential structure of the task adaptation. Fine-tuning with higher ranks also bypasses the issue of setting a LoRA rank lower than the intrinsic task dimensionality before fine-tuning which can yield suboptimal performance for more complex tasks; intrinsic task dimensionality can be encountered during fine-tuning or post-hoc as proposed by \method.

\noindent \textbf{Connections to Overparameterization and the Lottery Ticket Hypothesis.}
By training with a higher source rank $r_{src} > r_{tgt}$, we essentially perform the optimization in a less constrained, overparameterized space. This larger search space provides more degrees of freedom, making it easier to navigate the complex loss landscape and find a high-quality solution that effectively captures the low-dimensional manifold corresponding to the task. This principle may be seen as analogous to the findings of \cite{frankle2019lottery}, termed the Lottery Ticket Hypothesis, where an overparameterized network provides a richer substrate from which an efficient, high-performing subnetwork can be identified.

In our case, RSVD serves as a principled, data-driven, and efficient pruning mechanism; SVD guarantees the best low-rank approximation of a matrix with respect to the Frobenius norm. By retaining the top $r_{tgt}$ singular values and their corresponding singular vectors, we are preserving the directions of greatest variance in the learned delta matrix $\Delta W_{src}$. This effectively isolates the most significant, principal components of the task adaptation while filtering out potential noise or less impactful components that may have been learned due to overparameterization. 

\section{Additional Experiments}
\label{app:additional}
\begin{figure}[t!] 
    \centering 
        \begin{subfigure}[h]{0.32\textwidth}
        \centering
        \includegraphics[width=0.99\linewidth]{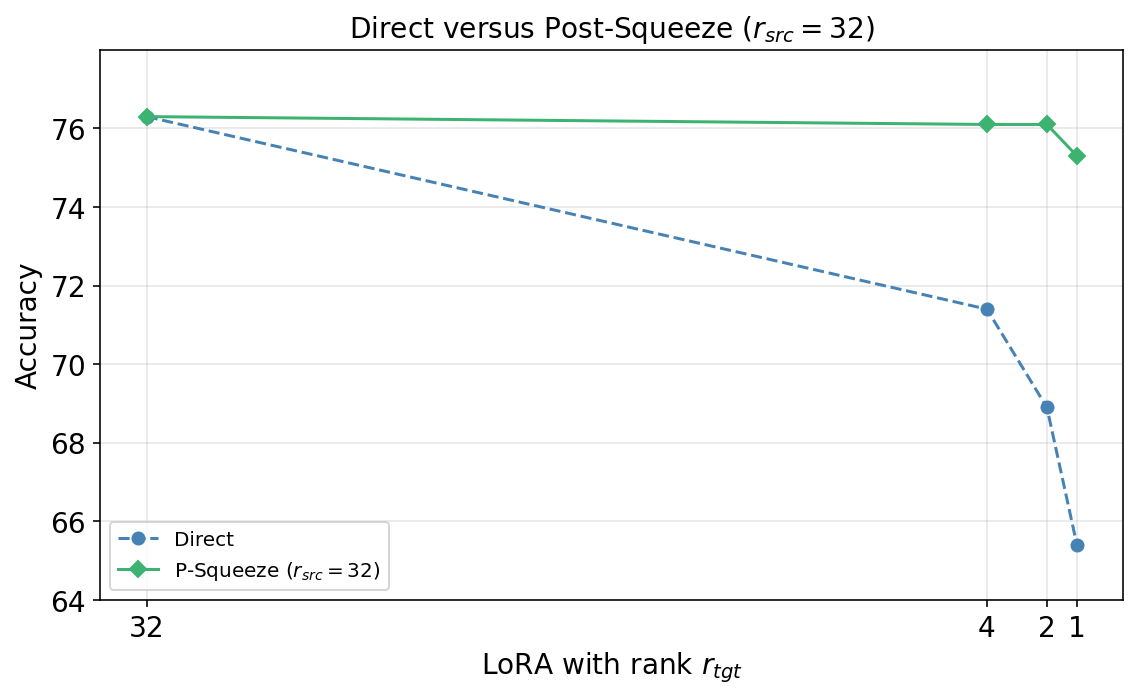}
        \caption{OK-VQA}
        \label{fig:subopt_okvqa}
    \end{subfigure}
    \begin{subfigure}[h]{0.32\textwidth}
        \centering
        \includegraphics[width=0.99\linewidth]{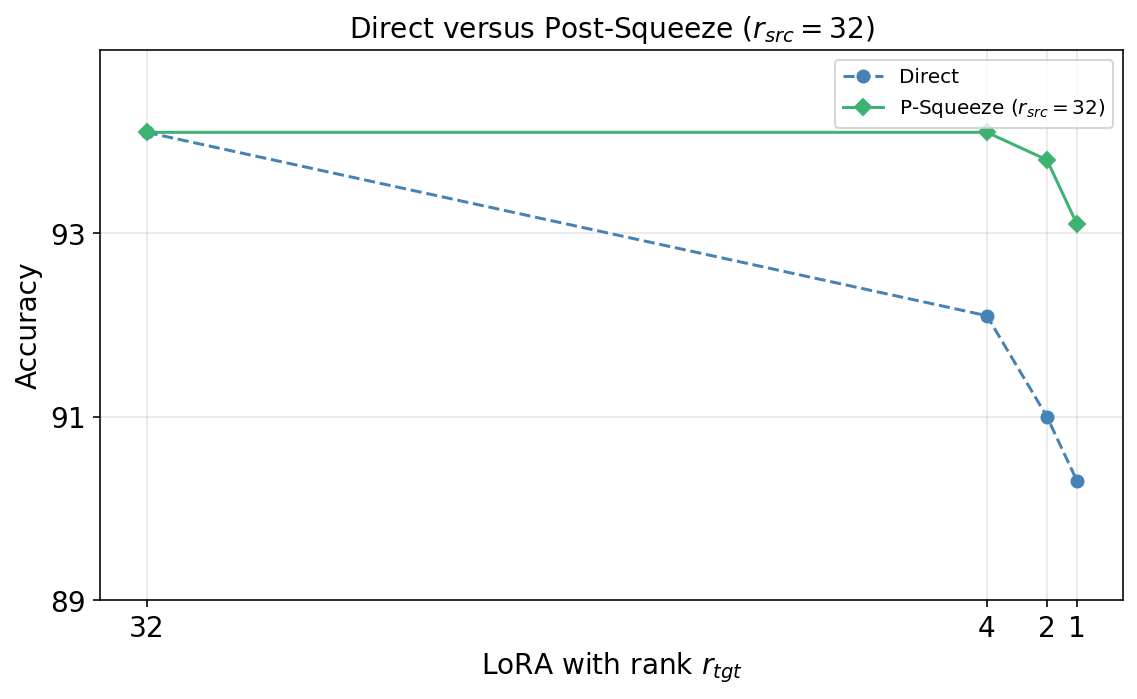}
        \caption{TallyQA}
        \label{fig:subopt_tallyqa}
    \end{subfigure}

    \caption{Performance over 2 VL tasks when we do hyperparameter search for the learning rate or LoRA-s only for the highest rank in the figures ($r_{src}=32$, higher ranks already saturate performance), and keep the same learning rate for direct fine-tuning at all the other (lower) ranks. A simple offline \textit{Post-Squeeze} method can bypass the hyperparameter search and yield better-performing LoRA-s without any fine-tuning at the lower ranks. See Figure~\ref{fig:suopts} in the main paper for similar patterns observed for text-based tasks.
    }
    \label{fig:subopts_vl}
\end{figure}

For the sake of brevity and clarity of presentation in the main paper, we have relegated additional experiments and analyses to this appendix. These results substantiate the work's primary claims and are consistent with the trends presented in the main paper body. 
\
\noindent \textbf{Figure~~\ref{fig:subopts_vl}} shows creation of lower-rank LoRA-s via \textit{Post-Squeeze} for 2 VL tasks when hyperparameter optimization is conducted only for the higher, source rank; the model is Gemma 3 4B IT. 

\begin{figure}[t!]
    \centering
    \includegraphics[width=0.48\columnwidth]{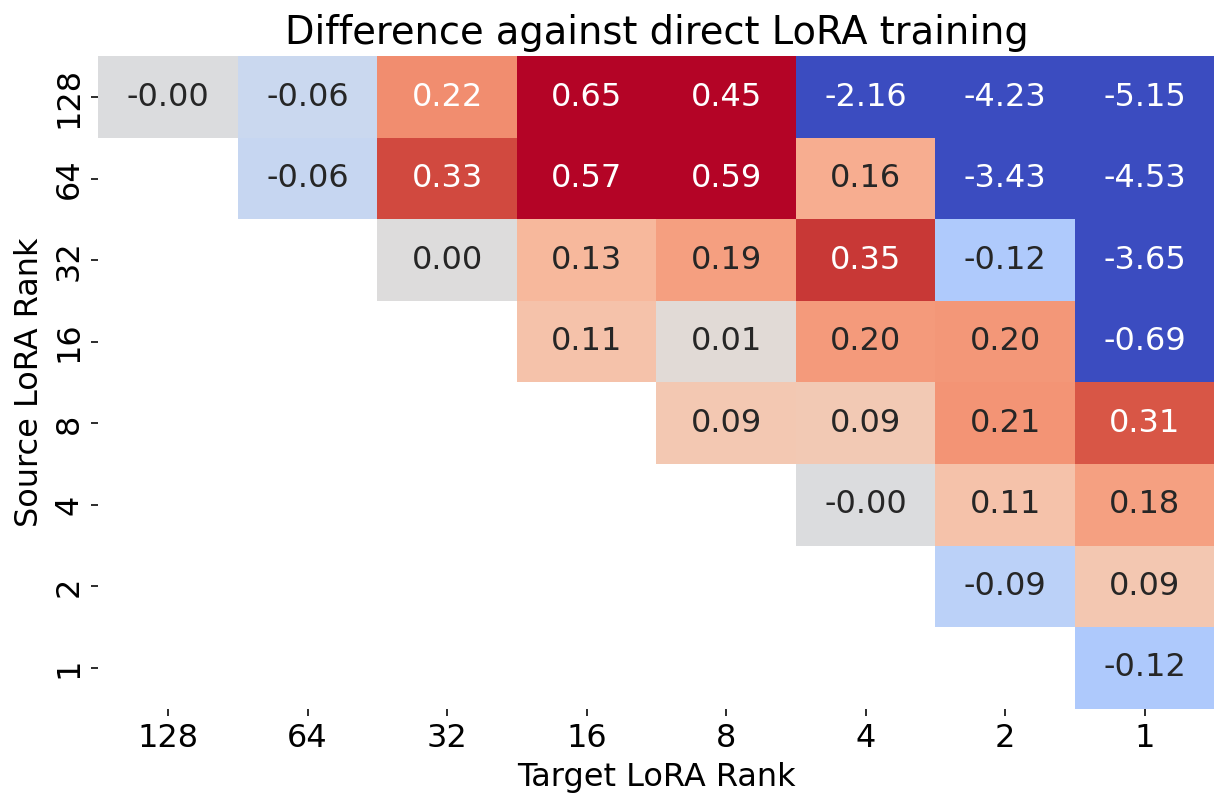}
    \caption{Performance difference heatmap for the Gemma 3 4B IT model averaged over the 10 VL tasks.}
    \label{fig:heatmap_4b_vl}
\end{figure}

\noindent \textbf{Figure~\ref{fig:heatmap_4b_vl}} shows averaged performance of various source-target rank configurations for \textit{Post-Squeeze} in the VL tasks; the model is Gemma 3 4B IT.

\begin{figure*}[t!] 
    \centering 
        \begin{subfigure}[h]{0.32\textwidth}
        \centering
        \includegraphics[width=0.99\linewidth]{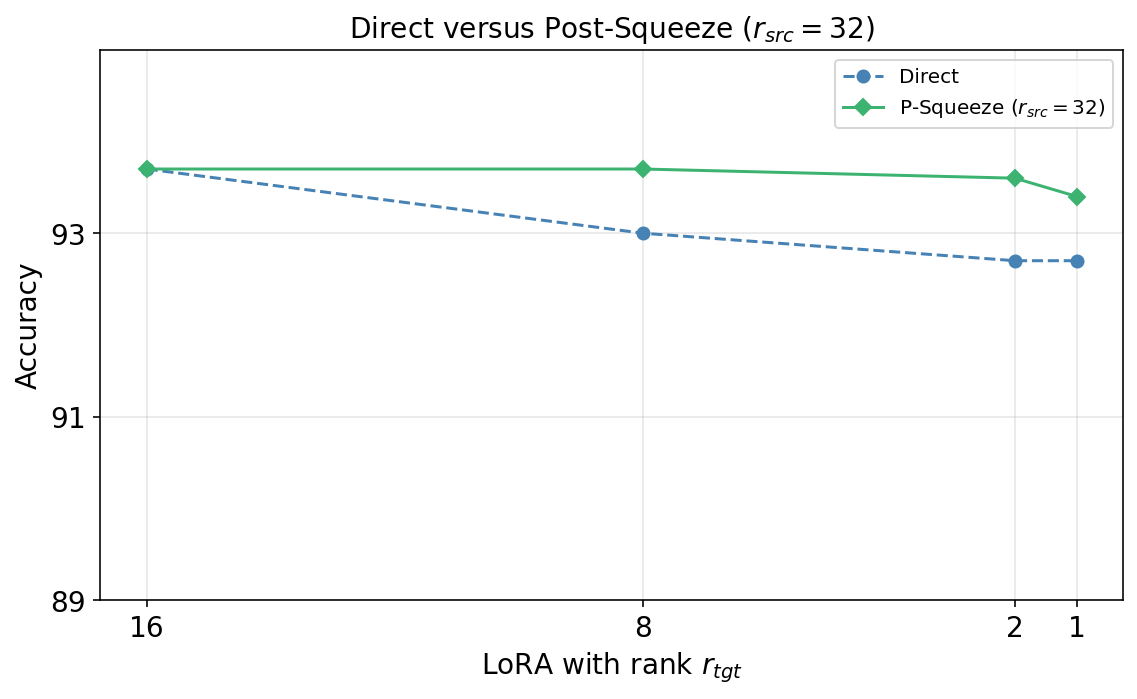}
        \caption{WNG-L}
        \label{fig:12b_wngl}
    \end{subfigure}
    \hfill 
    \begin{subfigure}[h]{0.32\textwidth}
        \centering
        \includegraphics[width=0.99\linewidth]{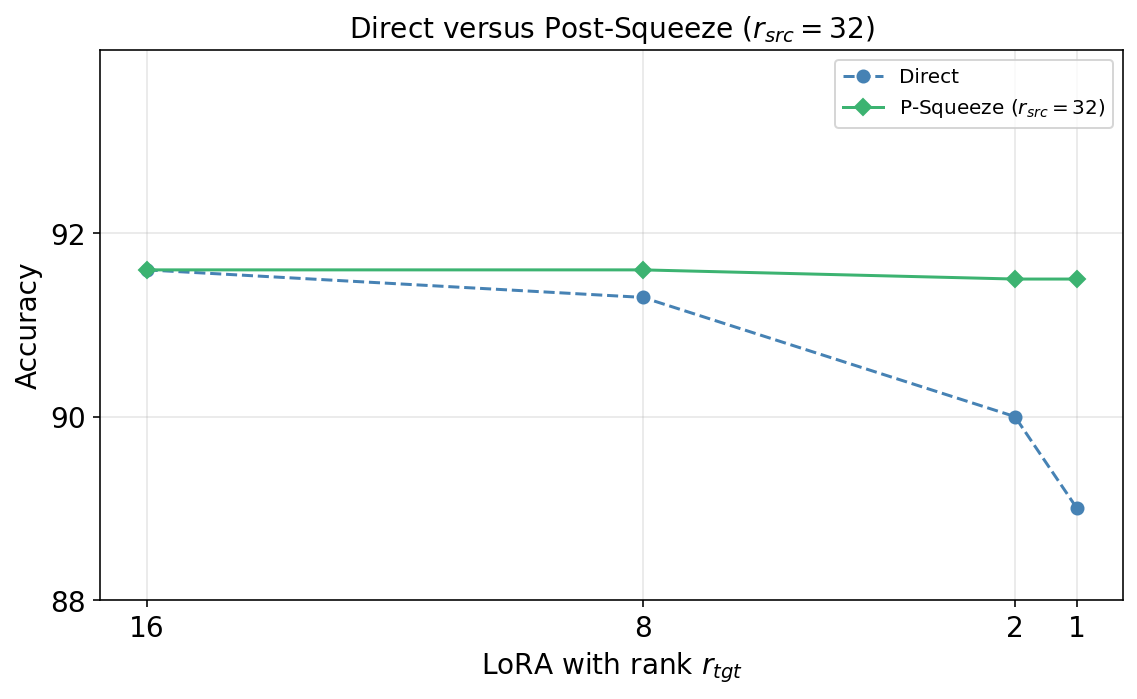}
        \caption{DROP}
        \label{fig:12b_drop}
    \end{subfigure}
    \hfill
    \begin{subfigure}[h]{0.32\textwidth}
        \centering
        \includegraphics[width=0.99\linewidth]{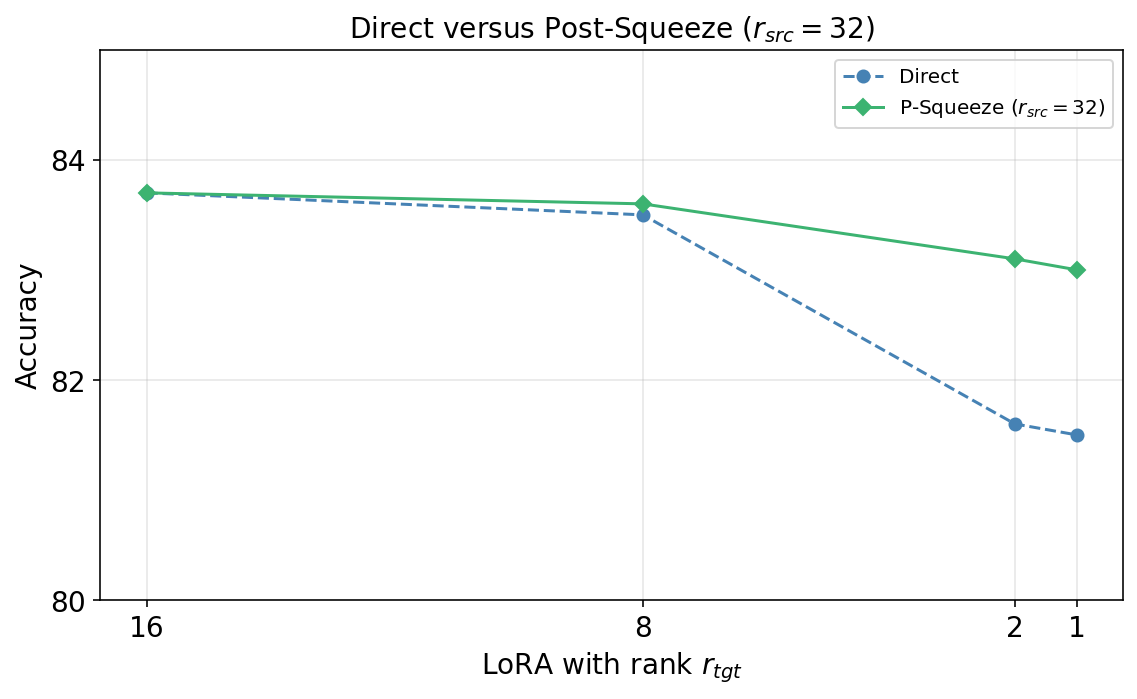}
        \caption{ANLI-r2}
        \label{fig:12b_anli}
    \end{subfigure}

    \caption{Performance over 3 representative text-based tasks with Gemma 12B IT as the base models. $r_{src}=16$.}
    \label{fig:12b}
\end{figure*}

\noindent \textbf{Figure~\ref{fig:12b}} shows a selection of results with \textit{Post-Squeeze} on Gemma 12B IT as the base model.

\noindent \textbf{Table~\ref{tab:post_direct}} shows per-task results of direct fine-tuning and \textit{Post-Squeeze} for two source-target rank combinations, where the averaged results are available as cell values in the heatmap in the main paper (Figure~\ref{fig:hmap_4b}); the model is Gemma 3 4B IT.

\noindent \textbf{Table~\ref{tab:finetuning_strategies_1b_text}} and \textbf{Table~\ref{tab:finetuning_strategies_4b_vl}} compare the per-task scores of direct fine-tuning with \textit{Cont-Squeeze} and \textit{In-Squeeze} for Gemma 1B IT on text-only tasks and for Gemma 4B IT on VL tasks, respectively. 

\noindent \textbf{Figure~\ref{fig:ret_tasks}} shows per-task variance retention scores over text-only tasks for three different source ranks $r_{src}$.

\captionsetup[subfigure]{margin={5mm,-2mm}}

\begin{table*}[t!]
\centering
\caption{Performance overview per single task (Accuracy \%) of \textit{Post-Squeeze} versus Direct Fine-Tuning on the Gemma 4B IT model across 13 text-only tasks for two source-target rank configurations. $\Delta$ indicates the absolute gain/loss of Post-Squeeze.}
\label{tab:post_direct}
\def\arraystretch{0.93}
\resizebox{0.78\textwidth}{!}{%
\begin{tabular}{@{}lcccccc@{}}
\toprule
 &  \multicolumn{3}{c}{\textbf{Target Rank = 4}} & \multicolumn{3}{c}{\textbf{Target Rank = 8}} \\
\cmidrule(lr){2-4} \cmidrule(lr){5-7}
\textbf{Task} & \textbf{Direct ($r=4$)} & \textbf{P-Squeeze (16$\rightarrow$4)} & \textbf{$\Delta$} & \textbf{Direct ($r=8$)} & \textbf{P-Squeeze (128$\rightarrow$8)} & \textbf{$\Delta$} \\ \midrule
WNG-L  & {91.07} & {90.99} & {-0.08} & 89.16 & 91.92 & +2.76 \\
BOOLQ  & {94.31} & {94.37} & {+0.06} & 94.11 & 94.28 & +0.17 \\
PIQA  & {93.05} & {92.72} & {-0.33} & 92.94 & 93.69 & +0.75 \\
DROP  & {88.51} & {89.63} & {+1.12} & 89.60 & 89.71 & +0.11 \\
ANLI-r2 & {80.13} & {81.08} & {+0.95} & 81.31 & 80.97 & -0.34 \\
PAWS & {97.45} & {97.28} & {-0.17} & 97.37 & 97.59 & +0.22 \\
HSWAG & {97.32} & {97.44} & {+0.12} & 97.42 & 97.58 & +0.16 \\
OBQA  & {94.71} & {95.19} & {+0.48} & 94.71 & 95.31 & +0.60 \\
GoE  & {82.81} & {83.38} & {+0.57} & 83.71 & 82.79 & -0.92 \\
ARC-E  & {95.01} & {94.94} & {-0.07} & 95.14 & 95.05 & -0.09 \\
ARC-C & {88.89} & {89.11} & {+0.22} &  88.98 & 88.76 & -0.22 \\
SIQA  & {89.74} & {89.25} & {-0.49} & 89.53 & 90.16 & +0.63 \\
MMLU  & {78.69} & {79.93} & {+1.24} & 79.92 & 80.36 & +0.44 \\ \midrule
\textbf{Average} & {90.13} & {\bf 90.41} & {\bf +0.28} & {90.30} & {\bf 90.63} & \textbf{+0.33} \\ \bottomrule
\end{tabular}%
}
\end{table*}

\begin{table*}[h!]
\centering
\caption{Performance comparison (Accuracy \%) of different fine-tuning strategies for Gemma 3 1B IT on text tasks. The best result in each row is highlighted in \textbf{bold}.}
\label{tab:finetuning_strategies_1b_text}
\resizebox{0.85\textwidth}{!}{
\begin{tabular}{@{}lcccccccc@{}}
\toprule
& \multicolumn{3}{c}{\textbf{Direct Fine-tuning ($r=1$)}} & \multicolumn{3}{c}{\textbf{Cont-Squeeze (128 $\rightarrow$ 1)}} & \multicolumn{2}{c}{\textbf{In-Squeeze (128 \ldots $\rightarrow$ 1)}} \\
\cmidrule(lr){2-4} \cmidrule(lr){5-7} \cmidrule(lr){8-9}
\textbf{Task} & \textbf{+0 steps} & \textbf{+200 steps} & \textbf{+700 steps} & \textbf{+0 steps} & \textbf{+200 steps} & \textbf{+700 steps} & \textbf{Standard} & \textbf{Min steps} \\ \midrule
WNG-L & 81.97 & 82.02 & 81.80 & 37.67 & 81.16 & 81.38 & \textbf{82.44} & 81.29 \\
BOOLQ & \textbf{91.26} & 91.00 & 90.79 & 51.30 & 90.81 & 90.56 & 90.29 & 90.92 \\
PIQA & 86.94 & 86.44 & 86.47 & 82.39 & 87.17 & 87.53 & \textbf{88.17} & 87.28 \\
DROP & \textbf{78.24} & 76.18 & 75.21 & 69.83 & 75.72 & 76.39 & 76.58 & 78.17 \\
ANLI-r2 & \textbf{75.11} & 74.72 & 72.38 & 21.26 & 73.66 & 72.88 & 74.16 & 74.33 \\
PAWS & 96.59 & 96.61 & 96.48 & 63.47 & 96.43 & \textbf{96.70} & 96.28 & 96.65 \\
HSWAG & 89.14 & 87.11 & 87.99 & 65.55 & 88.41 & 88.89 & 89.76 & \textbf{90.58} \\
OBQA & 87.74 & 87.50 & 88.10 & 87.74 & 88.22 & \textbf{89.06} & 87.26 & 88.70 \\
GoE & 82.51 & 81.84 & 81.28 & 80.23 & 82.16 & 81.94 & \textbf{83.15} & 83.17 \\
ARC-E & 85.20 & 85.07 & 85.39 & 35.94 & 85.50 & 85.89 & 85.35 & \textbf{86.00} \\
ARC-C & 74.70 & 75.48 & 75.13 & 43.01 & 74.13 & 74.18 & 73.61 & \textbf{74.78} \\
SIQA & 83.68 & 83.16 & 84.25 & 34.00 & 83.68 & 84.25 & 84.41 & \textbf{85.24} \\
MMLU  & 71.02 & 70.37 & 69.71 & 27.50 & 71.92 & 71.38 & 71.64 & \textbf{72.29} \\ \midrule
\textbf{Avg} & 83.39 & 82.88 & 82.69 & 53.84 & 83.00 & 83.16 & 83.32 & \textbf{83.80} \\ \bottomrule
\end{tabular}
}
\end{table*}

\begin{table*}[h!]
\centering
\caption{Performance comparison (Accuracy \%) of different fine-tuning strategies for Gemma 3 4B IT on vision-language tasks. \textit{0-S} refers to zero-shot performance of the base model without any task-specific
fine-tuning. The best result in each row is highlighted in \textbf{bold}.}
\label{tab:finetuning_strategies_4b_vl}
\resizebox{0.9\textwidth}{!}{
\begin{tabular}{@{}lccccccccc@{}}
\toprule
& {} & \multicolumn{3}{c}{\textbf{Direct Fine-tuning ($r=1$)}} & \multicolumn{3}{c}{\textbf{Cont-Squeeze (128 $\rightarrow$ 1)}} & \multicolumn{2}{c}{\textbf{In-Squeeze (128 \ldots $\rightarrow$ 1)}} \\
\cmidrule(lr){3-5} \cmidrule(lr){6-8} \cmidrule(lr){9-10}
\textbf{Task} &  \textbf{0-S} & \textbf{+0 steps} & \textbf{+200 steps} & \textbf{+700 steps} & \textbf{+0 steps} & \textbf{+200 steps} & \textbf{+700 steps} & \textbf{Standard} & \textbf{Min steps} \\ \midrule
ai2d & {\em 59.0} & 67.73 & 67.16 & 66.29 & 67.37 & 68.30 & 67.73 & \textbf{68.71} & 67.73 \\
aokvqa & {\em 89.0} &  91.94 & 91.33 & 91.27 & \textbf{92.68} & 92.44 & 92.37 & 92.00 & 91.70 \\
countbenchqa & {\em 87.4} & 91.84 & 91.47 & 91.94 & 91.20 & \textbf{92.59} & 92.49 & 92.22 & \textbf{92.59} \\
docvqa & {\em 80.3} &\textbf{89.63} & 89.52 & 89.44 & 77.86 & 89.44 & 89.22 & 89.58 & 89.52 \\
infovqa & {\em 56.2} & 81.56 & 81.07 & \textbf{81.61} & 67.01 & 81.36 & 80.92 & 81.45 & 81.45 \\
ocrvqa & {\em 65.3} & 83.14 & 83.06 & 83.66 & 82.73 & 83.82 & \textbf{84.86} & 83.79 & 83.93 \\
okvqa & {\em 52.9} & 73.06 & 73.35 & 72.35 & 73.71 & \textbf{75.29} & 73.88 & 74.94 & 73.71 \\
scienceqa & {\em 92.3} & 98.00 & 98.00 & 97.94 & 96.17 & 97.58 & \textbf{98.18} & 97.55 & 98.06 \\
tallyqa & {\em 78.9} & \textbf{94.09} & 93.53 & 94.06 & 72.50 & 93.56 & 93.78 & 93.75 & 94.03 \\
textvqa & {\em 73.3} & 80.96 & \textbf{81.04} & 80.63 & 79.27 & 80.72 & 79.71 & 80.84 & 80.31 \\ \midrule
\textbf{Avg} & {\em 73.5} & 85.20 & 84.95 & 84.92 & 80.05 & \textbf{85.51} & 85.31 & 85.48 & 85.30 \\ \bottomrule
\end{tabular}
}
\end{table*}